\documentclass[preprint,authoryear]{elsarticle}

\usepackage{amssymb}

\usepackage{multicol,setspace}
\usepackage{multirow}
\usepackage{xcolor}
\usepackage{framed,multirow}
\usepackage{amsmath}
\usepackage{blkarray}
\usepackage{placeins}
\usepackage{graphicx}
\usepackage{array}
\usepackage[font=small,labelfont=bf]{caption}
\usepackage[subrefformat=parens]{subcaption}
\usepackage[justification=centering]{caption}

\usepackage{latexsym}
\usepackage[ruled,vlined, algo2e]{algorithm2e}
\usepackage[utf8x]{inputenc}
\DeclareMathOperator*{\argmin}{arg\,min}

\usepackage{comment}
\journal{Neural Networks}
\begin{document}

\begin{frontmatter}
\title{Augmented Bilinear Network for Incremental Multi-Stock Time-Series Classification}

\author[1]{Mostafa Shabani\corref{correspondingauthor}}
\cortext[correspondingauthor]{Corresponding author}
\ead{mshabani@ece.au.dk}

\author[2]{Dat Thanh Tran}
\ead{thanh.tran@tuni.fi}

\author[2]{Juho Kanniainen}
\ead{juho.kanniainen@tuni.fi}

\author[1]{\\Alexandros Iosifidis}
\ead{ai@ece.au.dk}

\address[1]{Department of Electrical and Computer Engineering, Aarhus University, Denmark}
\address[2]{Department
of Computing Sciences, Tampere University Tampere, Finland}

\begin{abstract}
Deep Learning models have become dominant in tackling financial time-series analysis problems, overturning conventional machine learning and statistical methods. Most often, a model trained for one market or security cannot be directly applied to another market or security due to differences inherent in the market conditions. In addition, as the market evolves through time, it is necessary to update the existing models or train new ones when new data is made available. This scenario, which is inherent in most financial forecasting applications, naturally raises the following research question: \textit{How to efficiently adapt a pre-trained model to a new set of data while retaining performance on the old data, especially when the old data is not accessible?} In this paper, we propose a method to efficiently retain the knowledge available in a neural network pre-trained on a set of securities and adapt it to achieve high performance in new ones. In our method, the prior knowledge encoded in a pre-trained neural network is maintained by keeping existing connections fixed, and this knowledge is adjusted for the new securities by a set of augmented connections, which are optimized using the new data. The auxiliary connections are constrained to be of low rank. This not only allows us to rapidly optimize for the new task but also reduces the storage and run-time complexity during the deployment phase. The efficiency of our approach is empirically validated in the stock mid-price movement prediction problem using a large-scale limit order book dataset. Experimental results show that our approach enhances prediction performance as well as reduces the overall number of network parameters. 
\end{abstract}


\end{frontmatter}

\section{Introduction}\label{S:intro}
Deep Learning has become prominent in tackling challenges in financial time series analysis \citep{tsantekidis2017forecasting,Tran2017, Minh2018, Zhang2019, Passalis2020, passalis2019deep, sirignano2019deep, Dixon2017sequence, Tran2019, Tran2019a, sezer2020financial}. Since the inception of electronic trading systems, large amounts of trade data have become available and accessible to many people. With large-scale data, conventional approaches based on linear and non-linear models trained with convex optimization have become less efficient in terms of computational complexity as well as prediction performance. Significant improvements in computing hardware coupled with increasing amounts of large-scale datasets have enabled the research community to harness the power of deep neural networks combined with stochastic optimization. The shift towards an end-to-end Deep Learning paradigm not only allows us to tackle larger and more complex problems but also enables us to focus on more important aspects of the real-world problems, such as the feasibility and efficiency when applying learning models to real-world financial problems \citep{sezer2020financial}.

Although it is less demanding in terms of memory complexity compared to non-linear models trained with convex optimization, optimizing deep neural networks using stochastic optimization is still time-consuming, since in order to train a well-performing network often requires a high number of iterations of mini-batch based optimization. The high computational cost is also associated with high energy usage, which can significantly affect the profitability of a Deep Learning solution. On the microeconomic level, since the data throughput is large and the frequency of changes in the data distribution is high, especially in highly liquid markets, we are faced with a continuous influx of data. On the macroeconomic level, the market continuously evolves to accommodate different phases of the economy. These unique features of the financial market can easily make a data-driven system obsolete, requiring it to be frequently updated with recent data. The necessity of frequent updates coupled with the high cost of updates has made computational complexity a critical factor when utilizing deep neural networks in financial analysis or forecasting systems.

There have been different approaches to reduce computational complexity when training deep neural networks, such as designing novel low-complexity network architectures \citep{kiranyaz2017progressive, tran2019heterogeneous,tran2019learning, tran2020progressive, kiranyaz2020operational, heidari2020progressive}, replacing existing ones with their low-rank counterparts \citep{denton2014exploiting, jaderberg2014speeding, tran2018improving, huang2018ltnn, ruan2020edp}, or adapting the pre-trained models to new tasks, i.e., performing Transfer Learning (TL) \citep{shao2014transfer, yang2015learning, ding2016incomplete, ding2018deep, fons2020augmenting} or Domain Adaptation (DA) learning \citep{duan2012domain, wang2019domain, zhao2020review, hedegaard2021supervised}. Among these approaches, model adaptation is the most versatile since a method in this category is often architecture-agnostic, being complementary to other approaches. In financial analysis or forecast settings, the ability to reuse existing pre-trained models can play an important role. The need to update a system on a regular basis arises in many situations in finance \citep{yu2007online, cavalcante2015approach, wang2014online}. Instead of training a new model from scratch with the new data, an efficient adaptation method will allow us to quickly adjust the existing model using the new data with less computational burden. Not only can this approach reduce the operating costs, thereby increasing the profit of the system, but also create the flexibility of more frequent updates. Besides the computational efficiency, model re-usability might also improve the modeling performance of a system. For example, data might be abundant for highly liquid markets/securities but might be scarce for less liquid markets/securities. As a result, we might not have sufficient data of an illiquid stock to train a deep neural network from scratch without overfitting. By taking advantage of models pre-trained on other stocks, one can obtain better performance even with a small amount of data in new stocks.

In this paper, we consider the following research problem: given a new task $\mathcal{T}_{\textrm{new}}$ defined on financial time-series data and a neural network $\mathcal{N}_{\textrm{old}}$, which has been trained on previously collected financial time-series data to solve a task $\mathcal{T}_{\textrm{old}}$ that is relevant to $\mathcal{T}_{\textrm{new}}$, the objective is to efficiently generate a new network $\mathcal{N}_{\textrm{new}}$ based on $\mathcal{N}_{\textrm{old}}$ that generalizes well for the new task $\mathcal{T}_{\textrm{new}}$ without having access to the data defining $\mathcal{T}_{\textrm{old}}$, and without harming the performance of $\mathcal{N}_{\textrm{old}}$ in $\mathcal{T}_{\textrm{old}}$. The tasks are often expressed via their datasets, and we refer to the datasets of $\mathcal{T}_{\textrm{old}}$ and $\mathcal{T}_{\textrm{new}}$ as $\mathcal{D}_{\textrm{old}}$ and $\mathcal{D}_{\textrm{new}}$, respectively. 
We refer to the above-described problem as the \textit{Incremental Multi-Stock Time-Series Analysis} problem. 
While multiple time-series analysis tasks can be formulated in this context, we focus on the case where the $\mathcal{T}_{\textrm{old}}$ corresponds to a mid-price direction prediction task defined on a set of stocks. This task can be formulated as a time-series classification problem. Historic data of $\mathcal{T}_{\textrm{old}}$ forming $\mathcal{D}_{\textrm{old}}$ is used to train the Deep Learning model $\mathcal{N}_{\textrm{old}}$. Given a pre-trained $\mathcal{N}_{\textrm{old}}$ and historic data of a (set of) new stock(s) forming $\mathcal{D}_{\textrm{new}}$, we would like to exploit the knowledge encoded in $\mathcal{N}_{\textrm{old}}$ to effectively be able to predict the direction of mid-price movements of the stocks belonging to both $\mathcal{T}_{\textrm{old}}$ and the stock(s) defining the new task $\mathcal{T}_{\textrm{new}}$.

The research problem described above is highly relevant to stock market data analysis since one gets access to historic data of new stocks in different time periods, and historic data used to train existing models can be either absent or so big that merging data in $\mathcal{D}_{\textrm{old}}$ and $\mathcal{D}_{\textrm{new}}$ to train $\mathcal{D}_{\textrm{new}}$ can be impractical due to their large size. 
While TL and DA have been widely adopted to solve related problems, they are not well-suited for addressing the problem of interest in our study. This is due to that they either require the use of both $\mathcal{D}_{\textrm{old}}$ and $\mathcal{D}_{\textrm{new}}$ to \textit{adapt} the parameters of $\mathcal{N}_{\textrm{old}}$ to the new task, or they create a second model $\mathcal{N}_{\textrm{new}}$ the parameters of which are initialized to those of $\mathcal{N}_{\textrm{old}}$ and are further fine-tuned using $\mathcal{D}_{\textrm{new}}$, thus leading to an inefficient solution.

In this paper, we propose a method for performing network augmentation by learning auxiliary neural connections that are complementary to the existing ones. This allows the new model to preserve prior knowledge and rapidly adapt to the new tasks, leading to improvements in performance. By using a low-rank approximation for the auxiliary connections, our method obtains additional efficiency in terms of overall operational cost. We demonstrate our approach with the Temporal Attention-augmented Bilinear Layer (TABL) \citep{Tran2019a} network architecture, which achieves state-of-the-art performance in the stock mid-price direction prediction. In addition, we also demonstrate that the proposed approach can generalize to Convolutional Neural Networks (CNN), which is another type of neural network architecture that is widely used in financial time series analysis \citep{Zhang2019, tsantekidis2017forecasting}. Using a large-scale Limit Order Book (LOB) dataset, our empirical study shows that the proposed method can indeed improve both prediction performance and operational efficiency of TABL networks as well as CNN networks, compared to TL and DA approaches.

The contributions of the paper are summarized as follows:
\begin{itemize}
    \item We describe a new research problem which is motivated by situations frequently arising in financial time-series analysis problems;
    \item We propose a solution to the research problem based on the TABL network, the effectiveness of which is demonstrated through experiments on two frequently occurring scenarios;
    \item We show that the proposed approach can be extended to other neural network types by providing a solution based on CNN;
    \item We improve the time and space complexities of the optimization process of our solution by using low rank tensor representations.
\end{itemize}

The remainder of paper is organized as follows. Related works in financial time series analysis are briefly presented in Section \ref{sec:Related_works}. The proposed method is described in Section \ref{sec:Proposed_method}. The complexity analysis of the proposed method is provided in section \ref{sec:Complexity_Analysis}. In Section \ref{sec:Experiments}, we describe in detail our experimental protocol and present the empirical results. Section \ref{sec:Conclusions} concludes our work and discusses potential future research directions.

\section{Related work}\label{sec:Related_works}
While econometric models can provide certain statistical insights with great transparency  \citep{kanjamapornkul2017support, kanjamapornkul2016study}, deep neural networks following the end-to-end data-driven learning paradigm led to significant improvements on the performance of financial time series prediction tasks. In \citep{tsantekidis2017forecasting}, a methodology based on Convolutional Neural Networks (CNN) was proposed for mid-price direction prediction. Deep Learning models utilizing CNN to capture the spatial structure of the LOB and Long Short Term Memory (LSTM) units to capture time dependencies were proposed in \citep{Zhang2019,Tsantekidis2020}. A multilayer Perceptron was used in \cite{sirignano2019deep} in the form of a spatial neural network. Adaptive input normalization jointly optimizing the input data normalization in the form of a neural layer with the parameters of the Deep Learning used for classification was proposed in \citep{passalis2019deep, tran2021bilinearInputNormalization}.

\subsection{Model Re-usability}
To address model re-usability, TL enables exploiting information from a pre-trained model that was trained on a source dataset, referred to as the base model hereafter, to improve the performance of the target model being trained on a target dataset \citep{pan2009survey}. Domains of the source and target datasets can be different, however, they must be related to each other. In cases where the source and target domains are highly dissimilar, negative transfer issues can appear, i.e., transferring knowledge adversely affects the performance of the target model \citep{rosenstein2005transfer}. In \citep{fawaz2018transfer}, the Dynamic Time Warping method (DTW) \citep{berndt1994using} was used to find similarities between the source and target datasets to avoid the negative transfer issue in TL of deep CNNs for time series classification tasks. Autonomous transfer learning (ATL) was proposed in \citep{pratama2019atl} to produce a domain invariant network and handle the problem of concept drifts \citep{gama2014survey} both in the source and target domains.  A hybrid algorithm based on TL, i.e. the Online Sequential Extreme Learning Machine with Kernels (OS-ELMK), and ensemble learning for time series prediction was proposed in \citep{ye2018novel} to handle the problem of wide variability between old data (base dataset) and new data (target dataset). In \citep{zuo2016fuzzy}, a fuzzy regression TL method was proposed, using the Takagi-Sugeno fuzzy regression model to transfer knowledge from a source domain to a target domain. Data augmentation was used to improve the performance of TL for stock price direction prediction in \citep{fons2020augmenting}. A TL framework for predicting stock price movements that uses
relationships between stocks to construct effective input was proposed in \citep{nguyen2019novel}. In \citep{koshiyama2020quantnet}, a TL architecture using encoder-decoder was proposed to learn market-specific trading strategies. Online transfer learning \citep{zhao2014online,ge2013oms} tries to handle the transfer learning problem within an online learning process. In these types of problems, we use the knowledge from some source domains to improve the performance of an online learning task in the target domain. Incremental Learning (IL) \citep{ross2008incremental} is another approach that can be used when we are dealing with a data stream and the prediction performance is reduced due to changes in the feature space of the new task. These approaches try to update some characteristics of fixed network structure to handle the problem of the new task. In \citep{pratama2019automatic}, neural networks with dynamically evolved capacity (NADINE) are proposed, which can update the network structure and improve the prediction performance based on changes in the learning environment. 

DA is another model re-usability approach. DA methods are transductive TL approaches with the assumption that the distributions of the source dataset and target dataset are different \citep{wang2018deep}. In \citep{ganin2016domain}, a feature learning approach was proposed that provides domain-invariant features. In \citep{long2015learning}, a Deep Adaptation Network (DAN) architecture was proposed that uses maximum mean discrepancy (MMD) \citep{gretton2012kernel} to find a domain-invariant feature space. In \citep{hedegaard2021supervised}, it was shown that supervised DA can be seen as a two-view Graph Embedding. In \citep{mahardhika2019autonomous}, a method was proposed that combines DA techniques and drift handling mechanism to solve the multistream classification problem under multisource streams. In multistream classification \cite{mahardhika2019autonomous,renchunzi2022automatic}, we have two datasets of source and target stream data which come from the same domain. The source stream dataset consists of labeled data, while the target stream dataset is unlabeled. The task is to predict the class labels of the target stream data and address challenges related to infinite length and concept drift. In this type of problem, we have access to the data of both the source and target data. One of the major differences between TL and DA is that DA requires all the data of both the source and target domains. As the source and target models need to be trained jointly, it is memory-intensive since the parameters of both models need to be updated \citep{pan2009survey}. When there is a major difference between the distribution of the base and target datasets, the performance of the base model may deteriorate. Multi-domain learning \citep{dredze2010multi,bulat2020incremental} methods incorporate the properties of multi-task learning \citep{caruana1997multitask} and domain adaptation. In multi-domain learning, the goal is to handle the same problem for different domains. In \citep{senhaji2020not}, an adaptive method for multi-domain learning is proposed that reduced the required base model parameters based on the complexity of the different domains coming from image classification problems.

\subsection{Temporal Attention-augmented Bilinear Layer}
\label{sec:original_TABL}
We will demonstrate our model augmentation approach with an instantiation of the Temporal Attention-augmented Bilinear Layer (TABL) network \citep{Tran2019a}, which has been proposed as an efficient and effective neural network architecture for financial time-series classification. In the following, we describe the working mechanism of the TABL network architecture, providing necessary details to understand our method described in Section \ref{sec:Proposed_method}.

\begin{figure*}[t!]
\captionsetup{singlelinecheck = false, justification=justified}
\centering
        \includegraphics[width=0.7\textwidth]{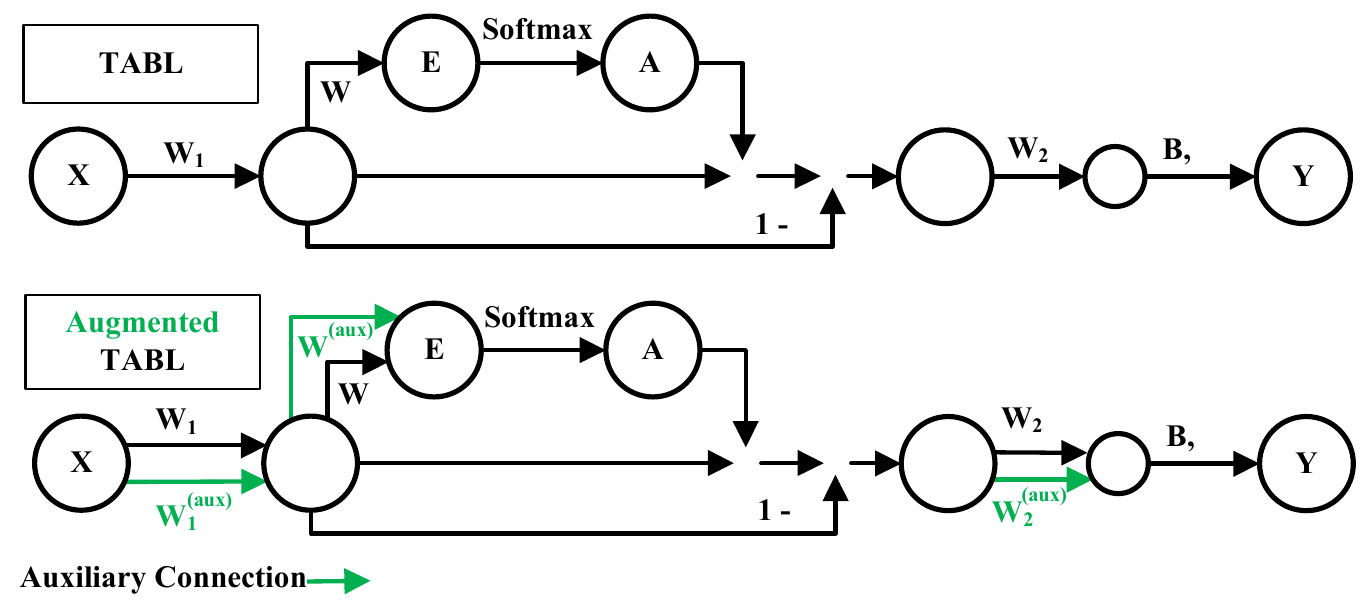}%
        \caption{Illustration of TABL (top) and the proposed Augmented TABL (bottom). Auxiliary connections are shown in green color.}
        \label{TABL_layer}
\end{figure*}

TABL combines the ideas of bilinear projection and attention mechanism. 
The input to a TABL is a multivariate time-series $\mathbf{X} \in \mathbb{R}^{D \times T}$, with $T$ denoting the number of instances combined in the temporal dimension to form the time-series and $D$ denoting the dimensionality of the instances forming the time-series. Given an input to the TABL, it generates an output multivariate time-series $\mathbf{Y} \in \mathbb{R}^{D' \times T'}$, where $T'$ and $D'$ denote the transformed number of instances and their dimensionality, respectively. This is performed through five computation steps: 
\begin{enumerate}
    \item A feature transformation of the input $\mathbf{X}$ is performed using a weight matrix $\mathbf{W}_1 \in \mathbb{R}^{D'\times D}$, producing the intermediate feature matrix $\bar{\mathbf{X}} \in \mathbb{R}^{D'\times T}$:
    \begin{equation}
        \bar{\mathbf{X}} = \mathbf{W}_1 \mathbf{X}. \label{eq:TABLa}
    \end{equation}
        
    \item The relative importance of the instances forming the time-series is computed by:
    \begin{equation}
        \mathbf{E} = \bar{\mathbf{X}} \mathbf{W}, \label{eq:TABLb}
    \end{equation}
    where $\mathbf{W} \in \mathbb{R}^{T\times T}$ is a structured matrix that has fixed diagonal elements equal to $1/T$. By learning non-diagonal elements, the matrix $\mathbf{W}$ expresses weights encoding the pair-wise instance importance in the transformed feature space $\mathbb{R}^{D'}$, while the self-importance of all instances are set to be equal.
    
    \item The importance scores stored at the elements $e_{ij}$ of $\mathbf{E}$ are normalized in a row-wise manner to produce an attention mask $\mathbf{A} \in \mathbb{R}^{D'\times T}$ formed by the attention scores. This is done using the soft-max function:
    \begin{equation}
        \alpha_{ij} = \frac{ exp(e_{ij}) }{ \sum_{k=1}^T exp(e_{ik}) }. \label{eq:TABLc}
    \end{equation}
    Each row of the attention matrix $\mathbf{A}$ sums up to 1, distributing the importance scores along the time dimension for each dimension of the transformed input time-series data.
    
    \item The attended features $\tilde{\mathbf{X}} \in \mathbb{R}^{D'\times T}$ are computed by applying the attention mask to $\bar{\mathbf{X}}$. To enable a soft attention, a learnable parameter $\lambda$ is used to weight the contribution of the attended $\bar{\mathbf{X}} \odot \mathbf{A}$ and the original features $\bar{\mathbf{X}}$:
    \begin{equation}
        \tilde{\mathbf{X}} = \lambda (\bar{\mathbf{X}} \odot \mathbf{A}) + (1-\lambda) \bar{\mathbf{X}}. \label{eq:TABLd}
    \end{equation}
    $\lambda$ is constrained to have a value between $[0, 1]$.
    
    \item The final attended features $\tilde{\mathbf{X}}$ are linearly transformed in the second tensor mode by weight matrix $\mathbf{W}_2 \in \mathbb{R}^{T \times T'}$, shifted by the bias $\mathbf{B} \in \mathbb{R}^{D' \times T'}$, and activated by a nonlinear element-wise activation function $\phi(\cdot)$, e.g., the Rectified Linear Unit (ReLU) function:
    \begin{equation}
        \mathbf{Y} = \phi\left( \tilde{\mathbf{X}} \mathbf{W}_2 + \mathbf{B} \right). \label{eq:TABLe}
    \end{equation}
    
\end{enumerate}
A schematic illustration of the above computation steps is provided in Figure \ref{TABL_layer}. 

A TABL network is created by combining multiple TABLs or by combining TABLs with Bilinear Layers (BL). A BL is a layer in which the temporal attention branch is not used, or equivalently the value of parameter $\lambda$ is set to zero.

\section{Proposed Method}\label{sec:Proposed_method}
The proposed method is based on two main ideas, i.e. model augmentation and low-rank approximation. Although the method presented in this Section is described in detail for the TABL network architecture, this approach can be easily generalized to other architectures, for example, the Convolutional Neural Network architecture as we will show later in this section.

Based on the problem definition described in the Introduction, we want to use the information encoded in a $\mathcal{N}_{\textrm{old}}$ that was trained on $\mathcal{D}_{\textrm{old}}$ to improve the prediction performance for $\mathcal{D}_{\textrm{new}}$ with the following restrictions:
\begin{enumerate}
    \item without using the data in $\mathcal{D}_{\textrm{old}}$, as this data may not be available or may be very big leading to very computationally costly training for $\mathcal{D}_{\textrm{new}}$,
    \item without harming the performance of $\mathcal{N}_{\textrm{old}}$ on the original task, i.e., $\mathcal{T}_{\textrm{old}}$,
    \item without increasing the memory and the computation requirements much for operating on both tasks $\mathcal{T}_{\textrm{old}}$ and $\mathcal{T}_{\textrm{new}}$, after $\mathcal{N}_{\textrm{new}}$ is trained and deployed.  
\end{enumerate}
This is achieved by augmenting the parameters $\mathbf{\Theta}_{\mathcal{N}_{\textrm{old}}}$ of $\mathcal{N}_{\textrm{old}}$ with auxiliary parameters $\mathbf{\Theta}_{\textrm{aux}}$. That is, $\mathcal{N}_{\textrm{new}}$ is constructed by adding additional connections to the pre-trained neural network $\mathcal{N}_{\textrm{old}}$. In order to retain the prior knowledge in $\mathcal{N}_{\textrm{old}}$, its parameters $\mathbf{\Theta}_{\mathcal{N}_{\textrm{old}}}$ are fixed and we only optimize the auxiliary parameters $\mathbf{\Theta}_{\textrm{aux}}$ to learn additional information needed to perform well on the stocks defining $\mathbf{T}_{\textrm{new}}$.

There are different strategies to incorporate auxiliary connections to a pre-trained model. One of the most common approaches in transfer learning is to add new hidden layers after the last hidden layer, thereby extending the network's depth and replacing the prediction layer, also known as the penultimate layer. Instead of modifying the pre-trained model's topology, i.e., the number of layers and the size of each layer, we propose to augment the pre-trained model with auxiliary connections that are parallel to the existing ones, thereby keeping the original architecture design unchanged. The intuition behind this approach is that the architectural choice of a pre-trained model is often validated and obtained by extensive experimentation process. It has been shown that many state-of-the-art network architectures such as ResNet-50 \citep{He_2016_CVPR} or DenseNet121 \citep{huang2017densely} are not specific to a dataset, but perform well in many similar problems. Thus, by respecting the architectural choices of the pre-trained network $\mathcal{N}_{\textrm{old}}$, we can avoid the time-consuming process of validating architectural choices for the auxiliary connections.


\subsection{Augmented TABL}
The proposed model augmentation method is illustrated in Figure \ref{TABL_layer}. As can be seen from this figure, we incorporate into the pre-trained TABL three auxiliary connections (depicted in green color), which are parameterized by $\mathbf{W}_1^{(\textrm{aux})}$, $\mathbf{W}^{(\textrm{aux})}$ and $\mathbf{W}_2^{(\textrm{aux})}$, respectively. The dimensions of auxiliary parameters are exactly the same as those that they are augmenting. Particularly, $\mathbf{W}_1^{(\textrm{aux})} \in \mathbb{R}^{D'\times D}$ is added to augment $\mathbf{W}_1$; $\mathbf{W}^{(\textrm{aux})} \in \mathbb{R}^{T\times T}$ is added to augment $\mathbf{W}$; and $\mathbf{W}_2^{(\textrm{aux})} \in \mathbb{R}^{T\times T'}$ is added to augment $\mathbf{W}_2$. The transformations produced by the augmented TABL are described by the following equations:
\begin{eqnarray}
\bar{\mathbf{X}} &=& \mathbf{W}_1 \mathbf{X} + \mathbf{W}_1^{(\textrm{aux})} \mathbf{X}, \label{eq:aTABLa} \\
\mathbf{E} &=& \bar{\mathbf{X}} \mathbf{W} + \bar{\mathbf{X}} \mathbf{W}^{(\textrm{aux})}, \label{eq:aTABLb} \\
\alpha_{ij} &=& \frac{ exp(e_{ij}) }{ \sum_{k=1}^T exp(e_{ik}) }, \label{eq:aTABLc} \\
\tilde{\mathbf{X}} &=& \lambda (\bar{\mathbf{X}} \odot \mathbf{A}) + (1-\lambda) \bar{\mathbf{X}}, \label{eq:aTABLd} \\
\mathbf{Y} &=& \phi\left( \tilde{\mathbf{X}} \mathbf{W}_2 + \tilde{\mathbf{X}} \mathbf{W}_2^{(\textrm{aux})} + \mathbf{B} \right). \label{eq:aTABLe}
\end{eqnarray}

In Eq. (\ref{eq:aTABLa}), the intermediate feature matrix that is formed using the $W_1$ of pre-trained model is added to the transformed features matrix that is produced using auxiliary parameter $\mathbf{W}_1^{(\textrm{aux})}$. The matrix $\mathbf{E}$ (Eq. (\ref{eq:aTABLb})) which shows the relative importance of instances is produced by summing up the output of Eq. (\ref{eq:aTABLa}) with both $\mathbf{W}$ and $\mathbf{W}^{(\textrm{aux})}$. In the Eq. (\ref{eq:aTABLe}), the final attended features that are produced in Eq. (\ref{eq:aTABLd}) are multiplied by both $\mathbf{W}_2$ and $\mathbf{W}_2^{(\textrm{aux})}$ and added together. The output of this summation is shifted by $\mathbf{B}$ and activated by an activation function. As can be seen from Eqs. (\ref{eq:aTABLa})-(\ref{eq:aTABLe}), for each computation step that involves a weight matrix, we retain the transformations used by the pre-trained TABL, while learning additional (complementary) transformations through the auxiliary parameters ($\mathbf{W}_1^{(\textrm{aux})}$, $\mathbf{W}^{(\textrm{aux})}$, $\mathbf{W}_2^{(\textrm{aux})}$). By making the auxiliary computation steps mimic those in the original TABL, we respect the architectural design of the pre-trained model. In addition, since only the auxiliary parameters are updated during the optimization process, intermediate knowledge of the pre-trained model (preserved in its weight matrices) is still retained and we only learn auxiliary information to improve performance in the new task at hand.
\subsection{Low Rank Augmented TABL}
Besides model augmentation, our method exploits the idea of low-rank approximation. That is, we enforce a constraint (an upper-bound $K$) to the rank of each auxiliary weight matrix by representing it as a two-factor decomposition. Specifically, we define the auxiliary weight matrices as:
\begin{eqnarray}
\mathbf{W}_1^{(\textrm{aux})} &=& \mathbf{L}_1 \mathbf{R}_1, \label{eqLR1} \\
\mathbf{W}^{(\textrm{aux})} &=& \mathbf{L} \mathbf{R}, \label{eqLR2} \\
\mathbf{W}_2^{(\textrm{aux})} &=& \mathbf{L}_2 \mathbf{R}_2, \label{eqLR3}
\end{eqnarray}
where $\mathbf{L}_1 \in \mathbb{R}^{D' \times K}, \mathbf{R}_1 \in \mathbb{R}^{K \times D}, \mathbf{L} \in \mathbb{R}^{T \times K}, \mathbf{R} \in \mathbb{R}^{K \times T}, \mathbf{L}_2 \in \mathbb{R}^{T \times K}, \mathbf{R}_2 \in \mathbb{R}^{K \times T'}$, with $K \le \textrm{min}(D, D', T, T')$ being a hyperparameter value.

By constraining the value of $K$ to a small number, we enforce the weight matrices $\mathbf{W}_1^{(\textrm{aux})}$, $\mathbf{W}^{(\textrm{aux})}$ and $\mathbf{W}_2^{(\textrm{aux})}$ to have low ranks. The advantage of the low-rank approximation is two-fold. The first advantage is the reduction in computational and memory complexities during the optimization process. This will be analyzed in detail in the next subsection. In addition to improvements in complexity, the use of low-rank approximation can also have a regularization effect, thus improving the learning performance of the neural network. This is because low-rank approximation reduces the degrees of freedom in a given transformation, i.e., the number of parameters to be estimated, thus, it potentially reduces the overfitting effect when sufficient training data is not available.

To summarize, to take advantage of the pre-trained model to solve the new task $\mathcal{T}_{\textrm{new}}$ defined by the dataset $\mathcal{D}_{\textrm{new}}$, we solve the following optimization objective:
\begin{equation}
\argmin_{\substack{\mathbf{L}_1^{(l)}, \mathbf{L}^{(l)}, \mathbf{L}_2^{(l)}, \\
\mathbf{R}_1^{(l)}, \mathbf{R}^{(l)}, \mathbf{R}_2^{(l)} \\
l=1, \dots, L}} {} \frac{1}{N}\sum_{i=1}^{N} \mathcal{L}\left(\mathcal{N}_{\textrm{new}}(x_i, y_i) \right),
\end{equation}
where $(x_i, y_i) \in \mathcal{D}_{\textrm{new}}$ denotes the $i$-th training time-series and the corresponding target in the new dataset, $L$ denotes the number of TABLs in the new model, and $\mathcal{L}$ denotes the loss function.

\subsection{Complexity Analysis}\label{sec:Complexity_Analysis}
In this subsection, we provide our analysis on the complexity of the proposed method, as well as some notes on the implementation details. 
The first thing we should point out about the complexity of our method is that the inference complexity of the new model $\mathcal{N}_{\textrm{new}}$ is exactly the same as that of the pre-trained model $\mathcal{N}_{\textrm{old}}$. In other words, after the transfer learning process, the new model induces the same operating cost as the old model. This is because of the distributive property of matrix multiplications in Eq. (\ref{eq:aTABLa}), (\ref{eq:aTABLb}), (\ref{eq:aTABLe}). Let us denote:
\begin{equation}\label{eq_new_weights}
\begin{aligned}
\mathbf{W}_1^{(\textrm{new})} &= \mathbf{W}_1 + \mathbf{W}_1^{(\textrm{aux})}, \\
\mathbf{W}^{(\textrm{new})} &= \mathbf{W} + \mathbf{W}^{(\textrm{aux})}, \\
\mathbf{W}_2^{(\textrm{new})} &= \mathbf{W}_2 + \mathbf{W}_2^{(\textrm{aux})}.
\end{aligned}
\end{equation}
Eqs. (\ref{eq:aTABLa}), (\ref{eq:aTABLb}) and (\ref{eq:aTABLe}) become:
\begin{eqnarray}
\bar{\mathbf{X}} &=& \mathbf{W}_1^{(\textrm{new})} \mathbf{X}, \label{eq:bTABLa} \\
\mathbf{E} &=& \bar{\mathbf{X}} \mathbf{W}^{(\textrm{new})}, \label{eq:bTABLb} \\
\mathbf{Y} &=& \phi\left( \tilde{\mathbf{X}} \mathbf{W}_2^{(\textrm{new})} + \mathbf{B} \right). \label{eq:bTABLe}
\end{eqnarray}

After training, we can simply compute Eq. (\ref{eq_new_weights}) once and only store the values of $\mathbf{W}_1^{(\textrm{new})}, \mathbf{W}^{(\textrm{new})}, \mathbf{W}_2^{(\textrm{new})}$ for inference. In this way, we do not need to retain both the values of $\mathbf{W}_1, \mathbf{W}, \mathbf{W}_2$ of the pre-trained model and the values of $\mathbf{L}_1, \mathbf{R}_1, \mathbf{L}, \mathbf{R}, \mathbf{L}_2, \mathbf{R}_2$ of the auxiliary connections, which may consume more storage space than $\mathbf{W}_1^{(\textrm{new})}, \mathbf{W}^{(\textrm{new})}, \mathbf{W}_2^{(\textrm{new})}$. Since the matrix dimensions in Eqs. (\ref{eq:bTABLa})-(\ref{eq:bTABLe}) for the new model are exactly the same as those in Eqs. (\ref{eq:TABLa}), (\ref{eq:TABLb}) and (\ref{eq:TABLe}) for the pre-trained model, our method does not introduce any additional complexity during the inference/deployment phase.

The complexity during the training phase is dependent on the implementation details. Here we should note that there are two strategies to implement the proposed augmented TABL:

\begin{itemize}
    \item \textbf{Implementation Strategy 1 (IS1)}: during the forward pass, using Eqs. (\ref{eqLR1})-(\ref{eqLR3}), we first compute $\mathbf{W}_1^{(\textrm{aux})}$, $\mathbf{W}^{(\textrm{aux})}$ and $\mathbf{W}_2^{(\textrm{aux})}$ from $\mathbf{L}_1, \mathbf{R}_1$, $\mathbf{L}$, $\mathbf{R}$, $\mathbf{L}_2$ and $\mathbf{R}_2$. After that, using Eq. (\ref{eq_new_weights}), we can compute $\mathbf{W}_1^{(\textrm{new})}$, $\mathbf{W}^{(\textrm{new})}$, $\mathbf{W}_2^{(\textrm{new})}$, which are then used in Eqs. (\ref{eq:bTABLa})-(\ref{eq:bTABLe}) to compute the output of the augmented TABL.
    
    \item \textbf{Implementation Strategy 2 (IS2)}: in this case we do not compute explicitly the auxiliary weight matrices $\mathbf{W}_1^{(\textrm{aux})}$, $\mathbf{W}^{(\textrm{aux})}$ and $\mathbf{W}_2^{(\textrm{aux})}$ but we take advantage of the size of their low-rank approximations. Specifically, Eqs. (\ref{eq:aTABLa}), (\ref{eq:aTABLb}) and (\ref{eq:aTABLe}) are computed as follows, with the computation order from left to right and by respecting the priority of the parentheses:
    \begin{eqnarray}
    \bar{\mathbf{X}} &=& \mathbf{W}_1 \mathbf{X} + \mathbf{L}_1(\mathbf{R}_1 \mathbf{X}), \label{eq:cTABLa} \\
    \mathbf{E} &=& \bar{\mathbf{X}} \mathbf{W} + (\bar{\mathbf{X}}\mathbf{L}) \mathbf{R}, \label{eq:cTABLb} \\
    \mathbf{Y} &=& \phi\left( \tilde{\mathbf{X}} \mathbf{W}_2 + (\tilde{\mathbf{X}}\mathbf{L}_2) \mathbf{R}_2 + \mathbf{B} \right). \label{eq:cTABLe}
    \end{eqnarray}

\end{itemize}

Each of the above-mentioned strategies has its own advantages in terms of computational and memory complexity. Adopting the first implementation strategy (IS1) will lead to a faster forward-backward pass compared to the second strategy (IS2). This is because during stochastic optimization, we often update multiple samples in the same forward-backward pass, thus leading to $\mathbf{X}$ having another large dimension, which corresponds to the mini-batch size. Since each equation in IS2 involves two matrix multiplications with $\mathbf{X}$ or $\bar{\mathbf{X}}$ or $\tilde{\mathbf{X}}$, IS2 requires more calculations compared to IS1. The computational complexity estimates of IS1 and IS2 can be found in the Appendix. 

While adopting IS1 can lead to a faster forward-backward pass compared to IS2, this strategy also requires a higher amount of memory, which might not be feasible when training large networks on a GPU with limited memory. This is because using IS2, during the forward pass, we do not need to keep the intermediate outputs of $\mathbf{W}_1 \mathbf{X}$ in Eq. (\ref{eq:cTABLa}), $\bar{\mathbf{X}}\mathbf{W}$ in Eq. (\ref{eq:cTABLb}) and $\tilde{\mathbf{X}}\mathbf{W}_2$ in Eq. (\ref{eq:cTABLe}) for the gradient update computation in the backward pass. On the other hand, when using IS1 we need to keep all intermediate outputs in the forward pass in order to compute the gradient updates for the backward pass. In addition, since the auxiliary weight matrices are explicitly computed in the forward pass of IS1, we do not obtain any memory reduction from using the low-rank approximation as is the case for IS2.

\subsection{Augmented Convolution Layer}
As we mentioned in the beginning of this section, the proposed approach can be easily generalized to other architectures since most neural networks rely on linear or multilinear transformations. For Convolutional Neural Networks (CNNs), an augmented convolution layer can be formed by incorporating low-rank auxiliary filters to the pre-trained filters in a convolution layer. Since the filters in a convolution layer can be represented as a 3-mode (1D convolution layer) or 4-mode (2D convolution layer) tensor, the low-rank auxiliary filters can be represented in the Canonical Polyadic (CP) form \citep{kolda2009tensor}, similar to the low-rank CNN proposed in \citep{tran2018improving}. Let us denote $\mathcal{W}$ as the pre-trained convolution filters in a convolution layer, and $\mathbf{X}$ as the input time-series. Similar to an augmented TABL, The augmented convolution layer has the following form:

\begin{align}
    \mathbf{Y} &= \mathbf{X} \circledast (\mathcal{W} + \mathcal{W}^{(\textrm{aux})}) \\
               & = \mathbf{X} \circledast \mathcal{W} + \mathbf{X} \circledast \mathcal{W}^{(\textrm{aux})}),
\end{align}
where $\mathbf{Y}$ denotes the output of the convolution layer, $\circledast$ denotes the convolution operation, and $\mathcal{W}^{(\textrm{aux})}$ denotes the low-rank auxiliary filters. 

The convolutional architectures for time-series often consist of 1D convolution layers. Thus, $\mathcal{W}$ is a 3-mode tensor, i.e., $\mathcal{W} \in \mathbb{R}^{N \times D \times t}$ with $N$ denotes the number of filters, $D$ denotes the number of time-series and $t$ denotes the kernel size. Using the CP form, $\mathcal{W}^{(\textrm{aux})}$ can be represented as:

\begin{equation}
    \mathcal{W}^{(\textrm{aux})} = \sum_{k=1}^{K} \mathbf{w}_1(k) \otimes \mathbf{w}_2(k) \otimes \mathbf{w}_3(k),
\end{equation}
where $\mathbf{w}_1(k) \in \mathbb{R}^{N \times 1 \times 1}$, $\mathbf{w}_2(k) \in \mathbb{R}^{1 \times D \times 1}$, $\mathbf{w}_3(k) \in \mathbb{R}^{1 \times 1 \times t}$. $\otimes$ denotes the outer product and $K$ denotes the rank hyperparameter. 

\section{Experiments}\label{sec:Experiments}

\subsection{Dataset and Experimental Protocols}\label{sec:Experiments_dataset}
In order to evaluate the effectiveness of the proposed method in the research problem defined in Section \ref{S:intro}, we conducted experiments in the problem of stock mid-price direction prediction using the information appearing in Limit Order Books (LOB). Mid-price is the average value between the best bid (buy) price and the best ask (sell) price of a given stock. Although this quantity is just a virtual quantity, i.e., at a given time instance, no transaction can happen at the mid-price, the direction of mid-price change can capture the dynamics of a given stock, reflecting the supply and demand in the market. For this reason, mid-price direction prediction is a popular problem when analyzing LOB information \citep{Cont2011}.

The dataset used in our experiments is a public LOB dataset known as FI-2010 \citep{Ntakaris2017}, which consists of more than four million limit orders during the period of ten working days (from 1st of June to 14th of June 2010). The dataset contains limit orders for five companies traded in the Helsinki Exchange, operated by Nasdaq Nordic. At each time instance, the dataset provides the quotes of the top ten levels of the LOB, i.e., the top ten best bid prices and volumes and top ten best ask prices and volumes. The top quotes form $40$ different values corresponding to the bid and ask prices and volumes of the top $10$ LOB levels. Regarding the labels of the mid-price, at any given time instance, the database provides the direction (stationary, increasing, decreasing) of the mid-price after the next $H$ time instances, where $H \in \{10, 20, 30, 50, 100\}$ denotes the prediction horizon.

We followed a similar experimental protocol as in \citep{Tran2019a}: the input to all of the evaluated models was formed from the $10$ most recent limit order events, which consists of $10$ instances of $40$ dimensions standardized using z-score normalization. As has been shown in \citep{tran2021informative}, the adoption of information from all $10$ LOB levels is important for achieving high performance in learning-based methods. All models were trained to predict the mid-price direction after $10$ events, i.e., $H=10$. Regarding the train and test datasets, we used time-series of the first seven days for training and time-series of the last three days for testing. From the training set, we used the last $10\%$ of the time-series samples for validation purposes. In addition, we also evaluated our model augmentation strategy in an online learning setting, which is described in detail in Section \ref{sec:FV_experiments}. All models were optimized using the Adam optimizer with an initial learning rate of $0.01$. The learning rate was reduced whenever the validation loss reached a plateau.

Since FI-2010 is an imbalanced dataset with the majority of labels belonging to the stationary class, F1-score is used as the main performance metric, similar to prior works \citep{Tran2019a,Zhang2019,passalis2019deep}. The class distributions of 5 stocks individually and the overall class distributions for the dataset can be seen in Table~\ref{tab:class_dists}. In addition to F1-score, we also report average accuracy, precision and recall. Finally, we adopted the same weighted entropy loss function as in \citep{Tran2019a} to alleviate the effects of class imbalance:
\begin{equation}
    L= -\sum_{c=1}^{3}\frac{\beta}{N_{c}}y_{c} \log(\tilde{y}_{c}),
\end{equation}
where the $N_{c}$ is the number of samples in \emph{c}-th class, and the $y_{c}$ and $\tilde{y}_{c}$ are the true and predicted probabilities of \emph{c}-th class respectively. $\beta$ is a constant with a value of $1e6$. All experiments were run $5$ times and we report the mean and standard deviation of $5$ runs for each performance metric. Because of the random model parameter initialization and the stochastic nature of the Backpropagation algorithm, it is often the case that we obtain different network parameters, thus different performances at different runs. The use of the standard deviation of the performance achieved over different runs helps us evaluate how stable and reliable our model is given that it is trained using a stochastic optimization process. 

\begin{table}[ht!]
\captionsetup{singlelinecheck = false, justification=justified}
\caption{Distribution of classes in train and test dataset. Class 0 refers to the stock price remains the same, class 1 to the situation where the price goes up, and class 2 where the price goes downs.} 
\label{tab:class_dists} 
\centering
\resizebox{\textwidth}{!}{%
\begin{tabular}{|c|ccc|ccc|}
\hline
\multirow{2}{*}{Target stock} &
  \multicolumn{3}{c|}{Train dataset} &
  \multicolumn{3}{c|}{Test dataset} \\ \cline{2-7} 
 &
  \multicolumn{1}{c|}{Class 0} &
  \multicolumn{1}{c|}{Class 1} &
  Class 2 &
  \multicolumn{1}{c|}{Class 0} &
  \multicolumn{1}{c|}{Class 1} &
  Class 2 \\ \hline
Stock 1 &
  \multicolumn{1}{c|}{9013} &
  \multicolumn{1}{c|}{8198} &
  8060 &
  \multicolumn{1}{c|}{2108} &
  \multicolumn{1}{c|}{2507} &
  2440 \\ \hline
Stock 2 &
  \multicolumn{1}{c|}{28487} &
  \multicolumn{1}{c|}{11235} &
  11394 &
  \multicolumn{1}{c|}{16712} &
  \multicolumn{1}{c|}{4629} &
  4482 \\ \hline
Stock 3 &
  \multicolumn{1}{c|}{40489} &
  \multicolumn{1}{c|}{10228} &
  10452 &
  \multicolumn{1}{c|}{26067} &
  \multicolumn{1}{c|}{4136} &
  3974 \\ \hline
Stock 4 &
  \multicolumn{1}{c|}{22258} &
  \multicolumn{1}{c|}{9983} &
  10263 &
  \multicolumn{1}{c|}{17326} &
  \multicolumn{1}{c|}{4661} &
  4333 \\ \hline
Stock 5 &
  \multicolumn{1}{c|}{61116} &
  \multicolumn{1}{c|}{21991} &
  21790 &
  \multicolumn{1}{c|}{38829} &
  \multicolumn{1}{c|}{9458} &
  8756 \\ \hline
\multicolumn{1}{|l|}{All Stocks} &
  \multicolumn{1}{l|}{161363 (56\%)} &
  \multicolumn{1}{l|}{61635 (21\%)} &
  \multicolumn{1}{l|}{61959 (21\%)} &
  \multicolumn{1}{l|}{101042 (67\%)} &
  \multicolumn{1}{l|}{25391 (16\%)} &
  \multicolumn{1}{l|}{23985 (15\%)} \\ \hline
\end{tabular}%
}
\end{table}

\subsection{Experimental Setup 1} \label{Experimental_s1}
In order to simulate an experiment following the definition of the research problem defined in Section \ref{S:intro}, we conducted experiments for both TABL and CNN architectures with the following setup:
\begin{itemize}
    \item For each stock $S$ in the database, we consider the data belonging to $S$ as the new dataset $\mathcal{D}_{\textrm{new}}$ and the data belonging to the remaining four stocks as the old dataset $\mathcal{D}_{\textrm{old}}$.
    
    \item We use the notation \texttt{TABL-base} and \texttt{CNN-base} to denote the models that were trained on the $\mathcal{D}_{\textrm{old}}$ dataset. This is the pre-trained model $\mathcal{N}_{\textrm{old}}$ in our problem formulation. This model, when used directly to evaluate on the new dataset $\mathcal{D}_{\textrm{new}}$, can serve as a simple baseline to compare with our method.
    
    \item A TL approach taking advantage of the pre-trained model (\texttt{TABL-base} and \texttt{CNN-base}) and finetuning it on the new dataset $\mathcal{D}_{\textrm{new}}$ is also used as a baseline. We denote models following this TL approach as \texttt{TABL-fine-tune} and \texttt{CNN-fine-tune}.
    
    \item The augmented TABL models obtained by our method with two implementation strategies are denoted as \texttt{aTABL-IS1} and \texttt{aTABL-IS2}, respectively. Similarly, the augmented CNN model denoted as \texttt{aCNN}. Here we should note that, similar to \texttt{aTABL}, there are also two implementation strategies for \texttt{aCNN}. In our experiments, we simply used the second implementation strategy for \texttt{aCNN}. 
    
    \item In addition to the above, we also train a TABL model from scratch (random initialization) using both $\mathcal{D}_{\textrm{old}}$ and $\mathcal{D}_{\textrm{new}}$. This model, denoted as \texttt{TABL}, represents the scenario where we have access to both the old and new datasets at once. Similarly, \texttt{CNN} is used to denote the model that was trained using both $\mathcal{D}_{\textrm{old}}$ and $\mathcal{D}_{\textrm{new}}$. 
\end{itemize}

\begin{figure}[!h]
\captionsetup{singlelinecheck = false, justification=justified}
\centering
        \includegraphics[width=0.35\textwidth]{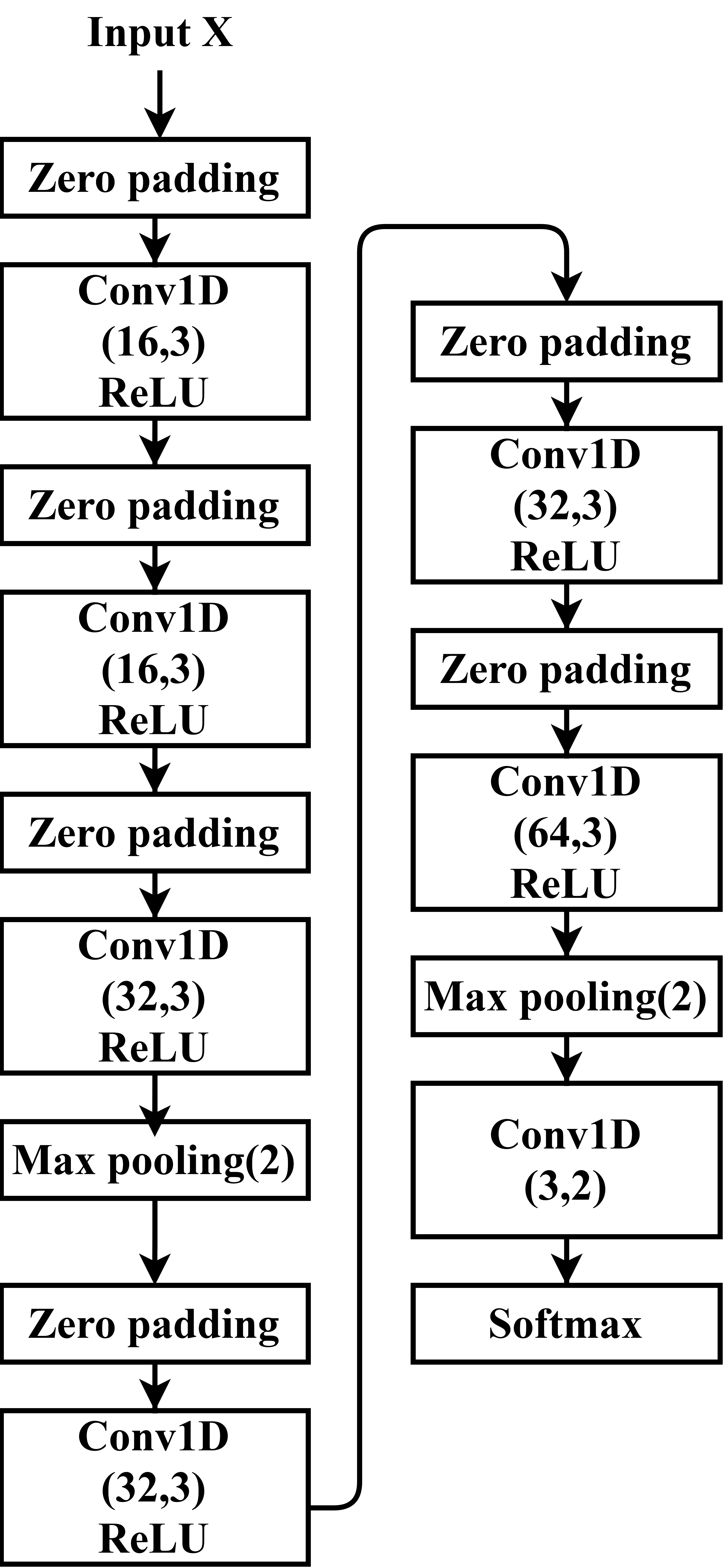}
        \caption{The CNN architecture with 7 convolution layers used in our experiments. Here the notation Conv1D (a, b) refers to a convolution layer having a filters, each of which has a kernel size of b. }
        \label{fig:fcn_diag}
\end{figure}

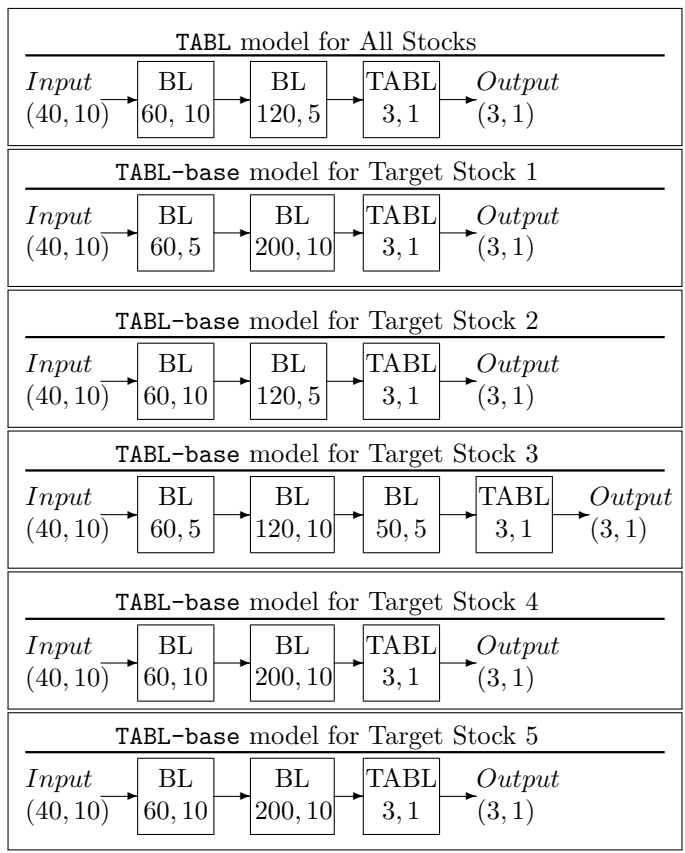
\begin{figure}[!h]
\captionsetup{singlelinecheck = false, justification=justified}
\setlength{\unitlength}{1cm} 
\begin{center}
\fbox{
\begin{picture}(8.5,1.6) 

\put(1.5,0){\framebox(1,1){\parbox{\unitlength}{\centering BL\\60, 10~}}}
\put(3,0){\framebox(1,1){\parbox{\unitlength}{\centering BL\\$120, 5$~}}}
\put(4.5,0){\framebox(1,1){\parbox{\unitlength}{\centering TABL\\$3, 1$~}}}
\put(1,0.5){\vector(1,0){0.5}}
\put(2.5,0.5){\vector(1,0){0.5}}
\put(4,0.5){\vector(1,0){0.5}}
\put(5.5,0.5){\vector(1,0){0.5}}

\put (0,0.4) {\begin{minipage}{\textwidth}$Input\\(40, 10)$\end{minipage}}
\put (6,0.4) {\begin{minipage}{\textwidth}$Output\\(3, 1)$\end{minipage}}
\put (2,1.2) {\begin{minipage}{\textwidth}\texttt{TABL} model for All Stocks
\end{minipage}}
\put (0,1) {\begin{minipage}{\textwidth}
\rule{8.5cm}{0.3mm}
\end{minipage}}
\end{picture}
}
\fbox{
\begin{picture}(8.5,1.6)(0,-0.1) 
 
\put(1.5,0){\framebox(1,1){\parbox{\unitlength}{\centering BL\\$60, 5$~}}}
\put(3,0){\framebox(1.1,1){\parbox{\unitlength}{\centering BL\\$200,10$~}}}
\put(4.5,0){\framebox(1,1){\parbox{\unitlength}{\centering TABL\\$3, 1$~}}}

\put(1,0.5){\vector(1,0){0.5}}
\put(2.5,0.5){\vector(1,0){0.5}}
\put(4.1,0.5){\vector(1,0){0.4}}
\put(5.5,0.5){\vector(1,0){0.5}}

\put (0,0.4) {\begin{minipage}{\textwidth}$Input\\(40, 10)$\end{minipage}}
\put (6,0.4) {\begin{minipage}{\textwidth}$Output\\(3, 1)$\end{minipage}}

\put (1.2,1.2) {\begin{minipage}{\textwidth}\texttt{TABL-base} model for Target Stock 1
\end{minipage}}
\put (0,1) {\begin{minipage}{\textwidth}
\rule{8.5cm}{0.3mm}
\end{minipage}}

\end{picture}
}
\fbox{
\begin{picture}(8.5,1.6) 
 
\put(1.5,0){\framebox(1,1){\parbox{\unitlength}{\centering BL\\$60, 10$~}}}
\put(3,0){\framebox(1,1){\parbox{\unitlength}{\centering BL\\$120, 5$~}}}
\put(4.5,0){\framebox(1,1){\parbox{\unitlength}{\centering TABL\\$3, 1$~}}}

\put(1,0.5){\vector(1,0){0.5}}
\put(2.5,0.5){\vector(1,0){0.5}}
\put(4,0.5){\vector(1,0){0.5}}
\put(5.5,0.5){\vector(1,0){0.5}}

\put (0,0.4) {\begin{minipage}{\textwidth}$Input\\(40, 10)$\end{minipage}}
\put (6,0.4) {\begin{minipage}{\textwidth}$Output\\(3, 1)$\end{minipage}}
\put (1.2,1.2) {\begin{minipage}{\textwidth}\texttt{TABL-base} model for Target Stock 2
\end{minipage}}
\put (0,1) {\begin{minipage}{\textwidth}
\rule{8.5cm}{0.3mm}
\end{minipage}}
\end{picture}
}
\fbox{
\begin{picture}(8.5,1.6)(0,-0.1) 
 
\put(1.5,0){\framebox(1,1){\parbox{\unitlength}{\centering BL\\$60, 5$~}}}
\put(3,0){\framebox(1.1,1){\parbox{\unitlength}{\centering BL\\$120,10$~}}}
\put(4.5,0){\framebox(1,1){\parbox{\unitlength}{\centering BL\\$50, 5$~}}}
\put(6,0){\framebox(1,1){\parbox{\unitlength}{\centering TABL\\$3, 1$~}}}

\put(1,0.5){\vector(1,0){0.5}}
\put(2.5,0.5){\vector(1,0){0.5}}
\put(4.1,0.5){\vector(1,0){0.4}}
\put(5.5,0.5){\vector(1,0){0.5}}
\put(7,0.5){\vector(1,0){0.5}}

\put (0,0.4) {\begin{minipage}{\textwidth}$Input\\(40, 10)$\end{minipage}}
\put (7.5,0.4) {\begin{minipage}{\textwidth}$Output\\(3, 1)$\end{minipage}}
\put (1.2,1.2) {\begin{minipage}{\textwidth}\texttt{TABL-base} model for Target Stock 3
\end{minipage}}
\put (0,1) {\begin{minipage}{\textwidth}
\rule{8.5cm}{0.3mm}
\end{minipage}}
\end{picture}
}
\fbox{
\begin{picture}(8.5,1.6) 
\put(1.5,0){\framebox(1,1){\parbox{\unitlength}{\centering BL\\$60, 10$~}}}
\put(3,0){\framebox(1.1,1){\parbox{\unitlength}{\centering BL\\$200, 10$~}}}
\put(4.5,0){\framebox(1,1){\parbox{\unitlength}{\centering TABL\\$3, 1$~}}}

\put(1,0.5){\vector(1,0){0.5}}
\put(2.5,0.5){\vector(1,0){0.5}}
\put(4.1,0.5){\vector(1,0){0.4}}
\put(5.5,0.5){\vector(1,0){0.5}}

\put (0,0.4) {\begin{minipage}{\textwidth}$Input\\(40, 10)$\end{minipage}}
\put (6,0.4) {\begin{minipage}{\textwidth}$Output\\(3, 1)$\end{minipage}}
\put (1.2,1.2) {\begin{minipage}{\textwidth}\texttt{TABL-base} model for Target Stock 4
\end{minipage}}
\put (0,1) {\begin{minipage}{\textwidth}
\rule{8.5cm}{0.3mm}
\end{minipage}}
\end{picture}
}
\fbox{
\begin{picture}(8.5,1.6)(0,-0.1) 
 
\put(1.5,0){\framebox(1,1){\parbox{\unitlength}{\centering BL\\$60, 10$~}}}
\put(3,0){\framebox(1.1,1){\parbox{\unitlength}{\centering BL\\$200,10$~}}}
\put(4.5,0){\framebox(1,1){\parbox{\unitlength}{\centering TABL\\$3, 1$~}}}
\put(1,0.5){\vector(1,0){0.5}}
\put(2.5,0.5){\vector(1,0){0.5}}
\put(4.1,0.5){\vector(1,0){0.4}}
\put(5.5,0.5){\vector(1,0){0.5}}
\put (0,0.4) {\begin{minipage}{\textwidth}$Input\\(40, 10)$\end{minipage}}
\put (6,0.4) {\begin{minipage}{\textwidth}$Output\\(3, 1)$\end{minipage}}
\put (1.2,1.2) {\begin{minipage}{\textwidth}\texttt{TABL-base} model for Target Stock 5
\end{minipage}}
\put (0,1) {\begin{minipage}{\textwidth}
\rule{8.5cm}{0.3mm}
\end{minipage}}
\end{picture}
}
\end{center}
\caption{The best base model network topology for each target stock. Each block corresponds to a layer (BL or TABL) and indicates the the output dimensions of the layer. The output with shape of (3,1) is a column vector with the 3 probability-like outputs of the network corresponding to the three classes. The top topology (\texttt{TABL}) refers to the case training is conducted using the training sets of all stocks.} 
\label{fig:Best_topoligoes} 
\end{figure}

To find the best network architecture for the pre-trained models (\texttt{TABL-base}), several configurations for hidden layers were validated. We followed the design in \citep{Tran2019a}, i.e., all hidden layers except the last one are Bilinear layers and the prediction layer is a TABL. Grid search was used for hyperparameter tuning. The results of ablation study for hidden layers can be seen in \ref{sec:Appendix_2}. The best network architectures for each pre-trained model corresponding to each experiment can be seen in Figure \ref{fig:Best_topoligoes}. For the CNN architecture, we adopted a conventional design pattern of 1D CNN for time-series, which is shown in Figure \ref{fig:fcn_diag}. 

\begin{table*}[h!]
\captionsetup{singlelinecheck = false, justification=justified}
\centering
\caption{Performance of TABL networks (Mean $\pm$ Standard Deviation) on the test sets of the Experimental Setup 1.}
\label{table:experiments_Result}
\resizebox{\linewidth}{!}{
\begin{tabular}{|l|l|c|c|c|c|c|}
\hline
\textbf{Target} & \textbf{Method} & \textbf{Accuracy (\%)} & \textbf{Precision (\%)} & \textbf{Recall (\%)} & \textbf{F1-score (\%)} & \textbf{Max Rank} \\ \hline
\multirow{5}{*}{\textbf{Stock 1}} & \texttt{TABL}      & 62.40 $\pm$ 00.10 & 62.10 $\pm$ 00.10 & 62.10 $\pm$ 00.10 & 62.08 $\pm$ 00.10 & - \\ \cline{2-7} 
                                  & \texttt{TABL-base}         & 63.96 $\pm$ 00.20 & 65.01 $\pm$ 00.30 & 64.45 $\pm$ 00.30 & 63.98 $\pm$ 00.20          & -          \\ \cline{2-7} 
                                  & \texttt{TABL-fine-tune}     & 64.97 $\pm$ 00.40         & 66.66 $\pm$ 00.50          & 65.73 $\pm$ 00.40        & 64.94 $\pm$ 00.40         & -          \\ \cline{2-7} 
                                  & \textbf{\texttt{aTABL-IS1}} & \textbf{65.37 $\pm$ 00.30} & \textbf{68.07 $\pm$ 00.50} & \textbf{66.34 $\pm$ 00.30} & \textbf{65.28 $\pm$ 00.30} & \textbf{9} \\ \cline{2-7} 
                                  & \textbf{\texttt{aTABL-IS2}} & \textbf{65.41 $\pm$ 00.40} & \textbf{68.68 $\pm$ 00.70} & \textbf{66.49 $\pm$ 00.40} & \textbf{65.28 $\pm$ 00.40} & \textbf{8} \\ \hline
\multirow{5}{*}{\textbf{Stock 2}} & \texttt{TABL}      & 78.44 $\pm$ 00.10 & 74.64 $\pm$ 00.20 & 66.26 $\pm$ 00.00 & 69.43 $\pm$ 00.00 & - \\ \cline{2-7} 
                                  & \texttt{TABL-base}          & 79.48 $\pm$ 00.40 & 77.21 $\pm$ 00.012 & 66.49 $\pm$ 00.10 & 70.31 $\pm$ 00.20          & -          \\ \cline{2-7} 
                                  & \texttt{TABL-fine-tune}     & 79.95 $\pm$ 01.00 & 79.97 $\pm$ 3.0 & 65.22 $\pm$ 00.10 & 69.91 $\pm$ 00.70          & -          \\ \cline{2-7} 
                                  & \textbf{\texttt{aTABL-IS1}} & \textbf{80.86 $\pm$ 00.30} & \textbf{82.62 $\pm$ 01.40} & \textbf{65.77 $\pm$ 00.20} & \textbf{70.98 $\pm$ 00.20} & \textbf{1} \\ \cline{2-7} 
                                  & \textbf{\texttt{aTABL-IS2}} & \textbf{81.07 $\pm$ 00.30} & \textbf{83.48 $\pm$ 01.00} & \textbf{65.72 $\pm$ 00.10} & \textbf{71.11 $\pm$ 00.20} & \textbf{4} \\ \hline
\multirow{5}{*}{\textbf{Stock 3}} & \texttt{TABL}      & 84.94 $\pm$ 00.10 & 75.65 $\pm$ 00.30 & 66.22 $\pm$ 00.10 & 70.03 $\pm$ 00.20 & - \\ \cline{2-7} 
                                  & \texttt{TABL-base}          & 82.96 $\pm$ 01.30 & 71.14 $\pm$ 02.50 & 66.17 $\pm$ 00.20 & 68.25 $\pm$ 01.20          & -          \\ \cline{2-7} 
                                  & \texttt{TABL-fine-tune}    & 84.61 $\pm$ 01.30 & 76.35 $\pm$ 04.10 & 64.79 $\pm$ 00.50 & 69.04 $\pm$ 01.10          & -          \\ \cline{2-7} 
                                  & \textbf{\texttt{aTABL-IS1}} & \textbf{85.82 $\pm$ 00.40} & \textbf{80.01 $\pm$ 01.90} & \textbf{65.26 $\pm$ 00.10} & \textbf{70.58 $\pm$ 00.50} & \textbf{4} \\ \cline{2-7} 
                                  & \textbf{\texttt{aTABL-IS2}} & \textbf{85.01 $\pm$ 01.90} & \textbf{77.74 $\pm$ 06.70} & \textbf{65.60 $\pm$ 00.30} & \textbf{70.02 $\pm$ 01.90} & \textbf{1} \\ \hline
\multirow{5}{*}{\textbf{Stock 4}} & \texttt{TABL}      & 77.55 $\pm$ 00.10 & 71.76 $\pm$ 00.20 & 64.25 $\pm$ 00.20 & 66.41 $\pm$ 00.10 & - \\ \cline{2-7} 
                                  & \texttt{TABL-base}         & 76.27 $\pm$ 03.20 & 69.47 $\pm$ 05.30 & 66.98 $\pm$ 00.60 & 67.74 $\pm$ 02.40          & -          \\ \cline{2-7} 
                                  & \texttt{TABL-fine-tune}     & 80.96 $\pm$ 00.40  & 80.66 $\pm$ 00.90  & 65.33 $\pm$ 00.80  & 70.20 $\pm$ 00.80            & -          \\ \cline{2-7} 
                                  & \textbf{\texttt{aTABL-IS1}} & \textbf{81.45 $\pm$ 00.30}  & \textbf{81.10 $\pm$ 01.30}  & \textbf{66.70 $\pm$ 00.30}  & \textbf{71.48 $\pm$ 00.20} & \textbf{1} \\ \cline{2-7} 
                                  & \textbf{\texttt{aTABL-IS2}} & \textbf{81.38 $\pm$ 00.50}  & \textbf{80.87 $\pm$ 01.60}  & \textbf{66.55 $\pm$ 00.30}  & \textbf{71.32 $\pm$ 00.40} & \textbf{1} \\ \hline
\multirow{5}{*}{\textbf{Stock 5}} & \texttt{TABL}      & 76.94 $\pm$ 00.10 & 70.03 $\pm$ 00.20 & 59.90 $\pm$ 00.20 & 63.39 $\pm$ 00.20 & - \\ \cline{2-7} 
                                  & \texttt{TABL-base}         & 67.72 $\pm$ 00.40 & 56.83 $\pm$ 00.70 & 59.82 $\pm$ 00.30 & 57.81 $\pm$ 00.10         & -          \\ \cline{2-7} 
                                  & \texttt{TABL-fine-tune}     & 77.35 $\pm$ 00.90  & 68.61 $\pm$ 01.50  & 64.57 $\pm$ 00.70  & 66.16 $\pm$ 01.20          & -          \\ \cline{2-7} 
                                  & \textbf{\texttt{aTABL-IS1}} & \textbf{78.77 $\pm$ 02.10}  & \textbf{72.24 $\pm$ 04.00}  & \textbf{64.43 $\pm$ 01.40}  & \textbf{67.18 $\pm$ 02.60} & \textbf{6} \\ \cline{2-7} 
                                  & \textbf{\texttt{aTABL-IS2}} & \textbf{78.50 $\pm$ 02.90}  & \textbf{71.53 $\pm$ 06.00}  & \textbf{63.59 $\pm$ 03.20}  & \textbf{66.47 $\pm$ 04.10} & \textbf{4} \\ \hline
\multirow{5}{*}{\textbf{Average}} & \texttt{TABL}      & 76.05 $\pm$ 08.30 & 70.84 $\pm$ 05.40 & 63.75 $\pm$ 02.70 & 66.27 $\pm$ 03.50 & - \\ \cline{2-7} 
                                  & \texttt{TABL-base}         & 74.08 $\pm$ 08.00 & 67.93 $\pm$ 07.60 & 64.78 $\pm$ 02.90 & 65.62 $\pm$ 04.90          & -          \\ \cline{2-7} 
                                  & \texttt{TABL-fine-tune}     & 77.57 $\pm$ 07.50 & 74.45 $\pm$ 06.50 & 65.13 $\pm$ 00.50 & 68.05 $\pm$ 02.40          & -          \\ \cline{2-7} 
                                  & \textbf{\texttt{aTABL-IS1}} & \textbf{78.45 $\pm$ 07.80} & \textbf{76.81 $\pm$ 06.30} & \textbf{65.70 $\pm$ 00.90} & \textbf{69.10 $\pm$ 02.70} & \textbf{-} \\ \cline{2-7} 
                                  & \textbf{\texttt{aTABL-IS2}} & \textbf{78.27 $\pm$ 07.60} & \textbf{76.46 $\pm$ 06.20} & \textbf{65.59 $\pm$ 01.20} & \textbf{68.84 $\pm$ 02.80} & \textbf{-} \\ \hline
\end{tabular}}
\end{table*}

\begin{table*}[ht!]
\captionsetup{singlelinecheck = false, justification=justified}
\centering
\caption{Performance of CNNs (Mean $\pm$ Standard Deviation) on the test sets of the Experimental Setup 1.}
\label{table:experiments_Result_FCN}
\resizebox{\linewidth}{!}{
\begin{tabular}{|l|l|l|l|l|l|l|}
\hline
\textbf{Target} & \textbf{Method} & \multicolumn{1}{c|}{\textbf{Accuracy (\%)}} & \multicolumn{1}{c|}{\textbf{Precision (\%)}} & \multicolumn{1}{c|}{\textbf{Recall (\%)}} & \multicolumn{1}{c|}{\textbf{F1-score (\%)}} & \textbf{Max Rank} \\ \hline
\multirow{4}{*}{Stock 1} & \texttt{CNN} & 53.93$\pm$0.013 & 53.69$\pm$0.014 & 53.05$\pm$0.012 & 52.45$\pm$0.013 & - \\ \cline{2-7} 
 & \texttt{CNN-base} & 52.94$\pm$0.005 & 52.4$\pm$0.005 & 51.78$\pm$0.005 & 50.78$\pm$0.006 & - \\ \cline{2-7} 
 & \texttt{CNN-fine-tune} & 55.07$\pm$0.0 & 54.99$\pm$0.005 & 54.67$\pm$0.005 & 54.52$\pm$0.006 & - \\ \cline{2-7} 
 & \textbf{\texttt{aCNN}} & \textbf{54.74$\pm$0.001} & \textbf{55.6$\pm$0.002} & \textbf{54.92$\pm$0.001} & \textbf{54.77$\pm$0.001} & \textbf{2} \\ \hline
\multirow{4}{*}{Stock 2} & \texttt{CNN} & 77.24$\pm$0.013 & 73.43$\pm$0.033 & 64.93$\pm$0.005 & 67.96$\pm$0.009 & - \\ \cline{2-7} 
 & \texttt{CNN-base} & 76.89$\pm$0.003 & 72.19$\pm$0.007 & 65.07$\pm$0.002 & 67.76$\pm$0.003 & - \\ \cline{2-7} 
 & \texttt{CNN-fine-tune} & 78.71$\pm$0.009 & 77.13$\pm$0.024 & 64.73$\pm$0.002 & 68.86$\pm$0.007 & - \\ \cline{2-7} 
 & \textbf{\texttt{aCNN}} & \textbf{78.48$\pm$0.008} & \textbf{75.73$\pm$0.022} & \textbf{65.37$\pm$0.001} & \textbf{69.01$\pm$0.006} & \textbf{1} \\ \hline
\multirow{4}{*}{Stock 3} & \texttt{CNN} & 82.93$\pm$0.013 & 71.77$\pm$0.034 & 64.92$\pm$0.006 & 67.67$\pm$0.01 & - \\ \cline{2-7} 
 & \textbf{\texttt{CNN-base}} & \textbf{83.03$\pm$0.01} & \textbf{71.64$\pm$0.025} & \textbf{64.99$\pm$0.002} & \textbf{67.77$\pm$0.009} & - \\ \cline{2-7} 
 & \texttt{CNN-fine-tune} & 76.45$\pm$0.042 & 64.18$\pm$0.029 & 61.4$\pm$0.034 & 60.97$\pm$0.042 & - \\ \cline{2-7} 
 & \texttt{aCNN} & 82.64$\pm$0.008 & 70.49$\pm$0.021 & 65.32$\pm$0.003 & 67.55$\pm$0.007 & \textbf{1} \\ \hline
\multirow{4}{*}{Stock 4} & \texttt{CNN} & 76.9$\pm$0.022 & 70.95$\pm$0.041 & 66.74$\pm$0.005 & 68.16$\pm$0.013 & - \\ \cline{2-7} 
 & \texttt{CNN-base} & 74.3$\pm$0.028 & 66.6$\pm$0.035 & 66.16$\pm$0.017 & 66.06$\pm$0.026 & - \\ \cline{2-7} 
 & \texttt{CNN-fine-tune} & 79.11$\pm$0.028 & 79.98$\pm$0.057 & 63.44$\pm$0.034 & 67.86$\pm$0.037 & - \\ \cline{2-7} 
 & \textbf{\texttt{aCNN}} & \textbf{79.42$\pm$0.0} & \textbf{74.93$\pm$0.002} & \textbf{67.17$\pm$0.0} & \textbf{70.13$\pm$0.001} & \textbf{2} \\ \hline
\multirow{4}{*}{Stock 5} & \texttt{CNN} & 67.34$\pm$0.032 & 56.72$\pm$0.014 & 56.75$\pm$0.019 & 56.19$\pm$0.003 & - \\ \cline{2-7} 
 & \texttt{CNN-base} & 67.89$\pm$0.024 & 56.29$\pm$0.022 & 56.24$\pm$0.012 & 56.06$\pm$0.016 & - \\ \cline{2-7} 
 & \texttt{CNN-fine-tune} & 67.52$\pm$0.055 & 60.78$\pm$0.034 & 53.0$\pm$0.021 & 52.69$\pm$0.025 & - \\ \cline{2-7} 
 & \textbf{\texttt{aCNN}} & \textbf{67.99$\pm$0.006} & \textbf{56.49$\pm$0.006} & \textbf{57.11$\pm$0.002} & \textbf{56.69$\pm$0.002} & \textbf{2} \\ \hline
\multirow{4}{*}{Average} & \texttt{CNN} & 72.15$\pm$0.11 & 66.27$\pm$0.089 & 61.78$\pm$0.057 & 63.19$\pm$0.071 & - \\ \cline{2-7} 
 & \texttt{CNN-base} & 71.82$\pm$0.105 & 64.56$\pm$0.085 & 61.23$\pm$0.058 & 62.24$\pm$0.072 & - \\ \cline{2-7} 
 & \texttt{CNN-fine-tune} & 72.63$\pm$0.089 & 68.56$\pm$0.102 & 59.59$\pm$0.054 & 61.3$\pm$0.074 & - \\ \cline{2-7} 
 & \textbf{\texttt{aCNN}} & \textbf{72.77$\pm$0.101} & \textbf{66.54$\pm$0.092} & \textbf{62.0$\pm$0.051} & \textbf{63.6$\pm$0.068} & \textbf{-}\\ \hline
\end{tabular}}
\end{table*}

Tables~\ref{table:experiments_Result} and~\ref{table:experiments_Result_FCN} provide the prediction performance of all models on the test sets defined on each experiment. The results are grouped based on the target stock $S$. At the bottom of Tables~\ref{table:experiments_Result} and~\ref{table:experiments_Result_FCN}, we report the average performance over the five target stocks. In addition, the last column shows the maximum rank value (chosen through validation) associated with our methods. As we have described two implementation strategies for the augmented TABL in Section \ref{sec:Proposed_method}, we also conducted the experiments and report the performance obtained by using these implementation strategies, denoted as \texttt{aTABL-IS1} and \texttt{aTABL-IS2} in Table~\ref{table:experiments_Result}.

From the experimental results in Tables \ref{table:experiments_Result} and \ref{table:experiments_Result_FCN}, we can clearly see that on average, the proposed model augmentation approach leads to performance improvements compared to the standard finetuning approach for both TABL and CNN network architectures. In addition, by adapting the pre-trained models on the new data using the proposed augmentation approach, we indeed observed performance improvements compared to the scenario when the pre-trained models are not adapted to the new data, i.e., \texttt{TABL-base} and \texttt{CNN-base}. This was not always the case for the standard finetuning approach since we observed performance degradation when comparing the performance of \texttt{CNN-fine-tune} and \texttt{CNN-base}. Regarding the two implementation strategies conducted for the TABL architectures, we can see that both strategies lead to very similar results. The differences stem from the stochastic nature of the optimizer as well as the initialization of the network parameters. Here we should note again that after training, the models obtained by our method have the same computational and memory complexities as the baseline or fine-tuned models. 

\begin{figure*}[t!]
\captionsetup{singlelinecheck = false, justification=justified}
\centering
        \includegraphics[width=\textwidth]{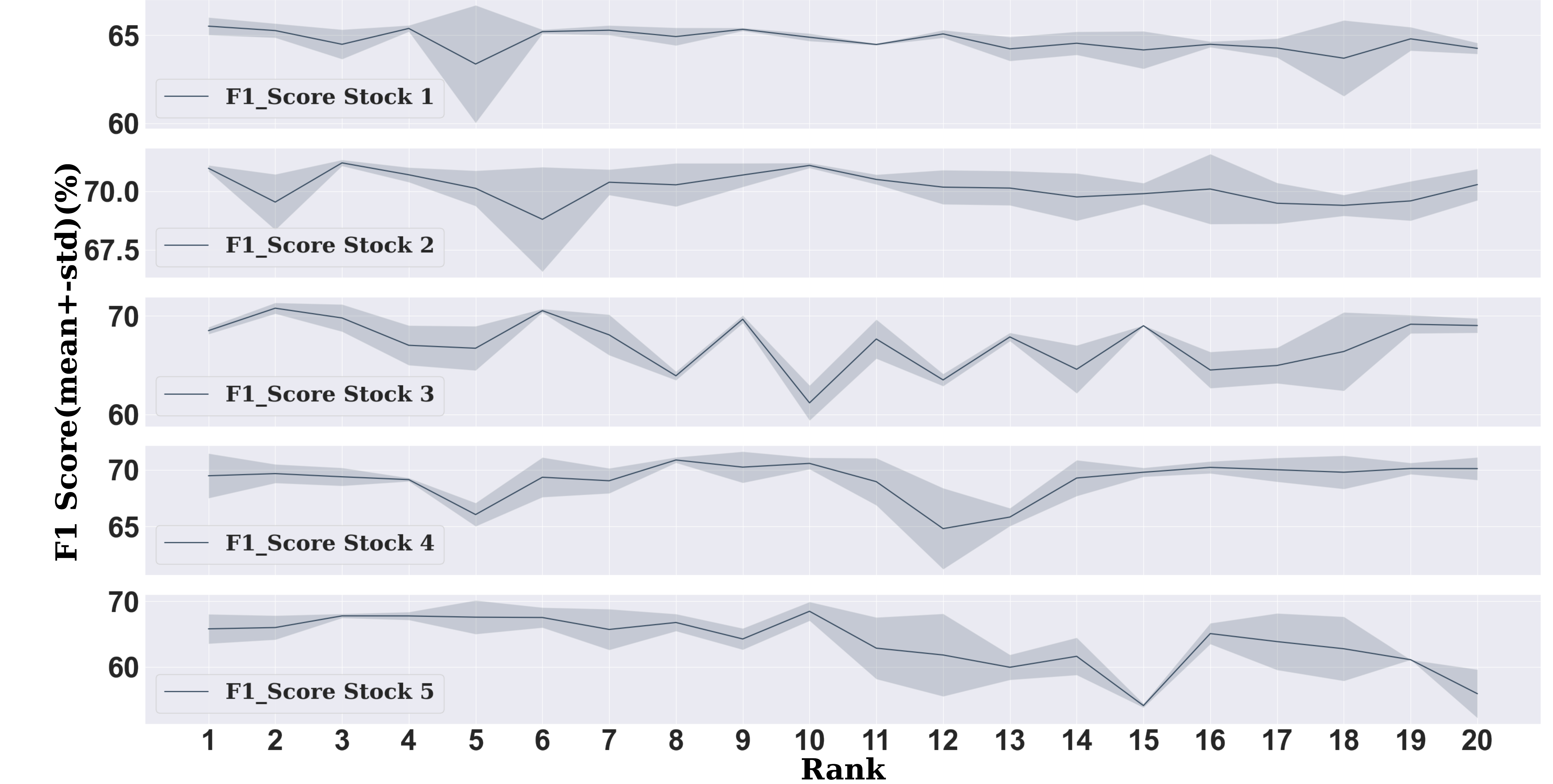}%
        \caption{Performance of \texttt{\textbf{aTABL-IS1}} for values of $K$~(rank of the model's weight matrices) between 1 to 20. The shadow of each line shows the standard deviation of results for each rank. }
        \label{fig:ranked_IS1}
\end{figure*}
\begin{figure*}[t!]
\captionsetup{singlelinecheck = false, justification=justified}
\centering
        \includegraphics[width=\textwidth]{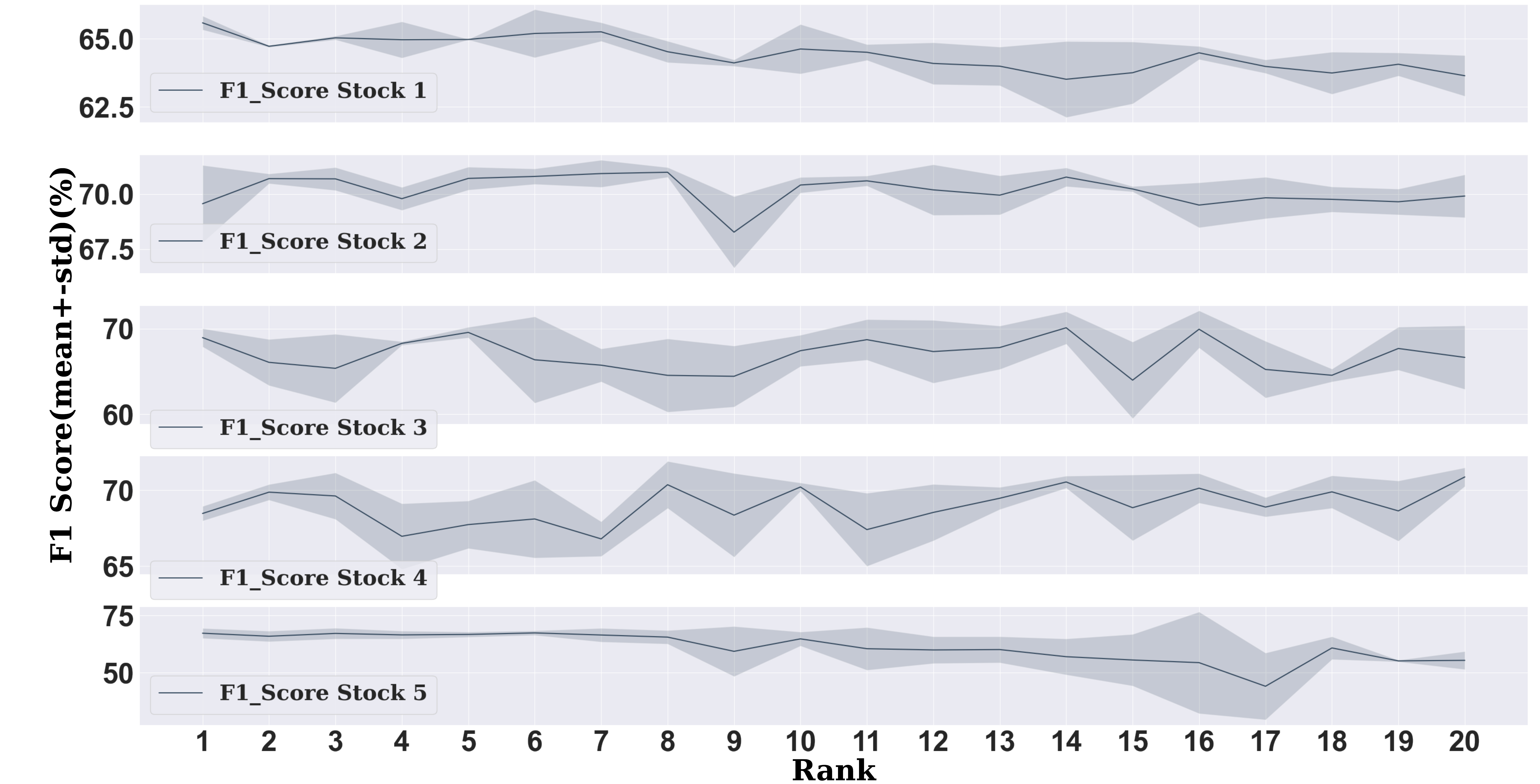}%
        \caption{Performance of \texttt{\textbf{aTABL-IS2}}  for values of $K$~(rank of the model's weight matrices) between 1 to 20. The shadow of each line shows the standard deviation of results for each rank.}
        \label{fig:ranked_IS2}
\end{figure*}

The rank value, which is a hyperparameter of the proposed augmentation method, can influence the efficiency of learning complementary knowledge from the new data. As mentioned before in this section, we chose the rank of the augmented models by running experiments on a range of different rank values and selected the one that produces the best F1-score on the training set. Figure \ref{fig:ranked_IS1} and Figure \ref{fig:ranked_IS2} show the effect of different ranks on the test performance of the proposed method. From these figures, it can be seen that there is no clear relation between the rank value and the generalization performance of the model. However, we can see that good performance can be obtained from low values of the rank.

\begin{figure}[!htb]
\captionsetup{singlelinecheck = false, justification=justified}
    \begin{minipage}[t]{.32\textwidth}
        \centering
        \includegraphics[width=1\textwidth]{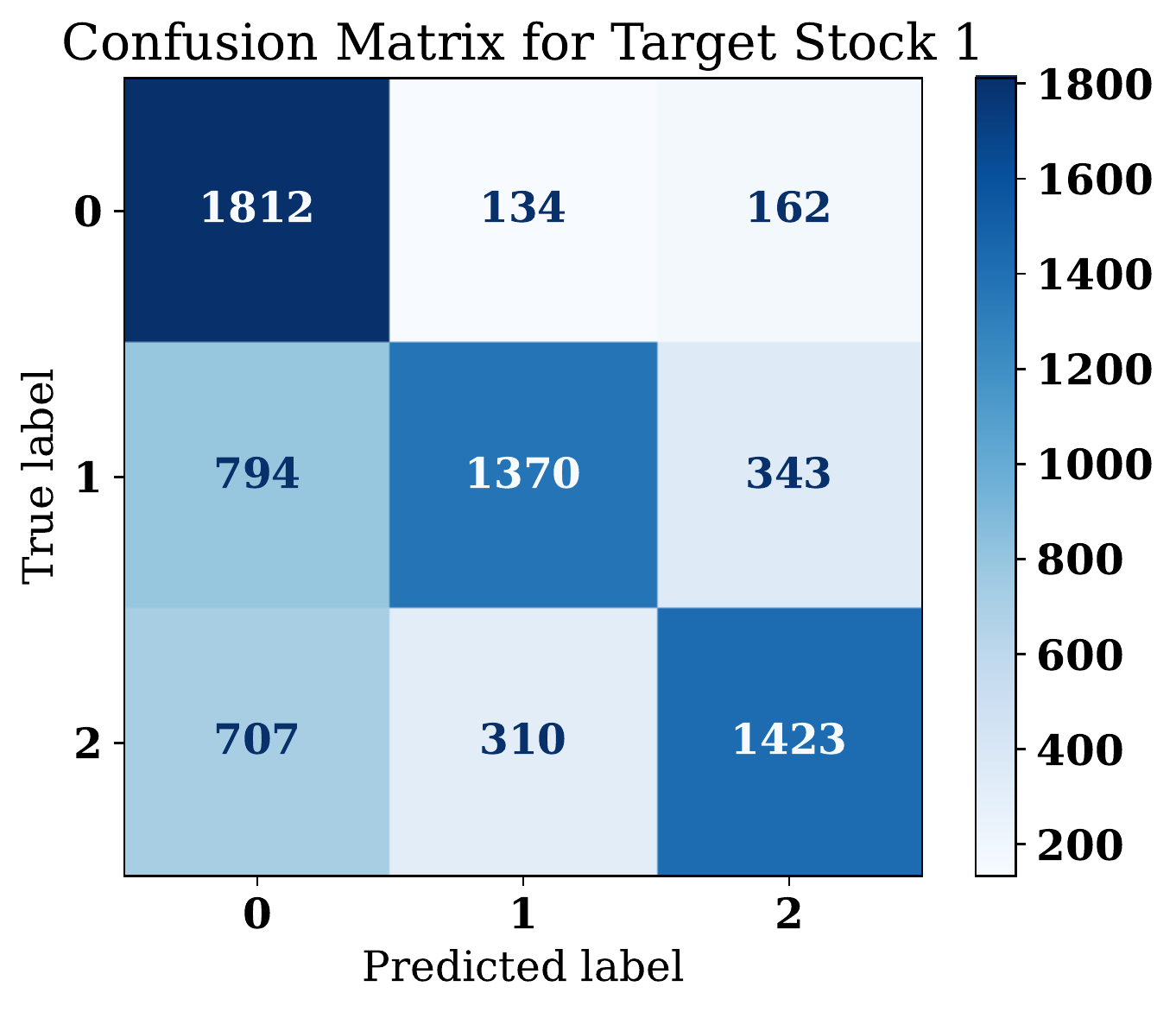}
    \end{minipage}
\hfill
    \begin{minipage}[t]{.32\textwidth}
        \centering
        \includegraphics[width=1\textwidth]{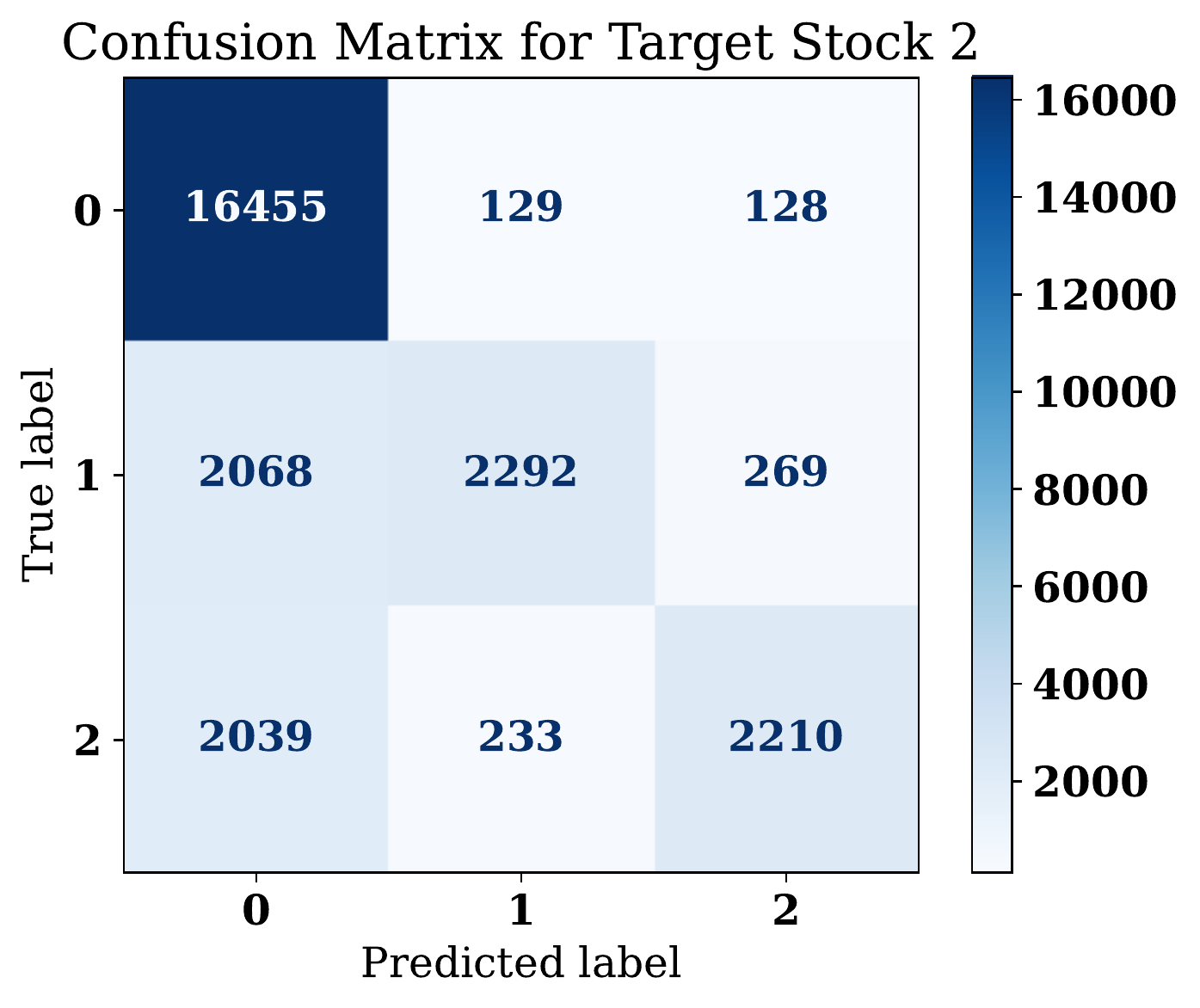}
        
    \end{minipage}
    \hfill
    \begin{minipage}[t]{.32\textwidth}
        \centering
        \includegraphics[width=1\textwidth]{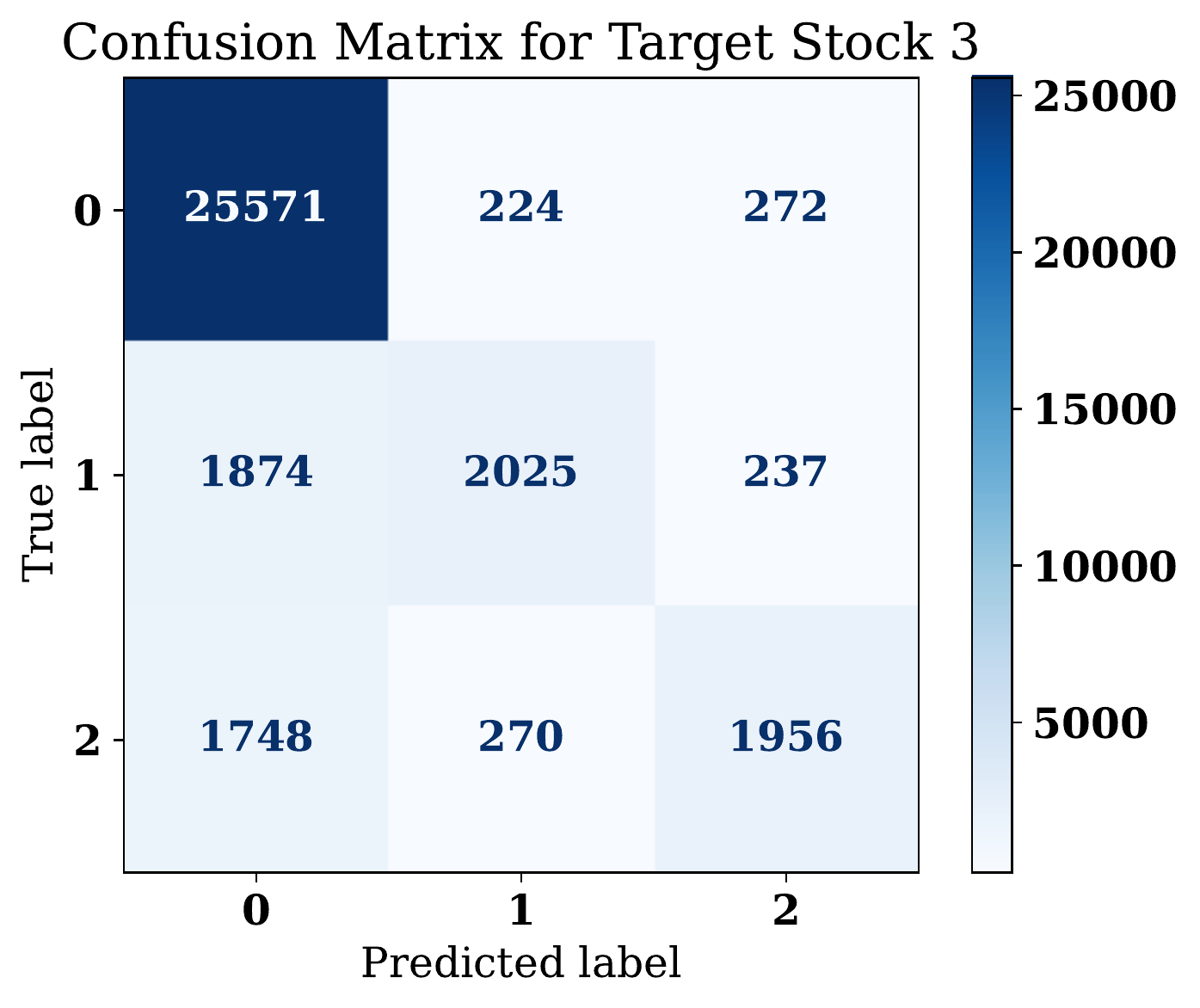}
        
    \end{minipage}
    \hfill
    \centering
    \begin{minipage}[t]{.32\textwidth}
        \centering
        \includegraphics[width=1\textwidth]{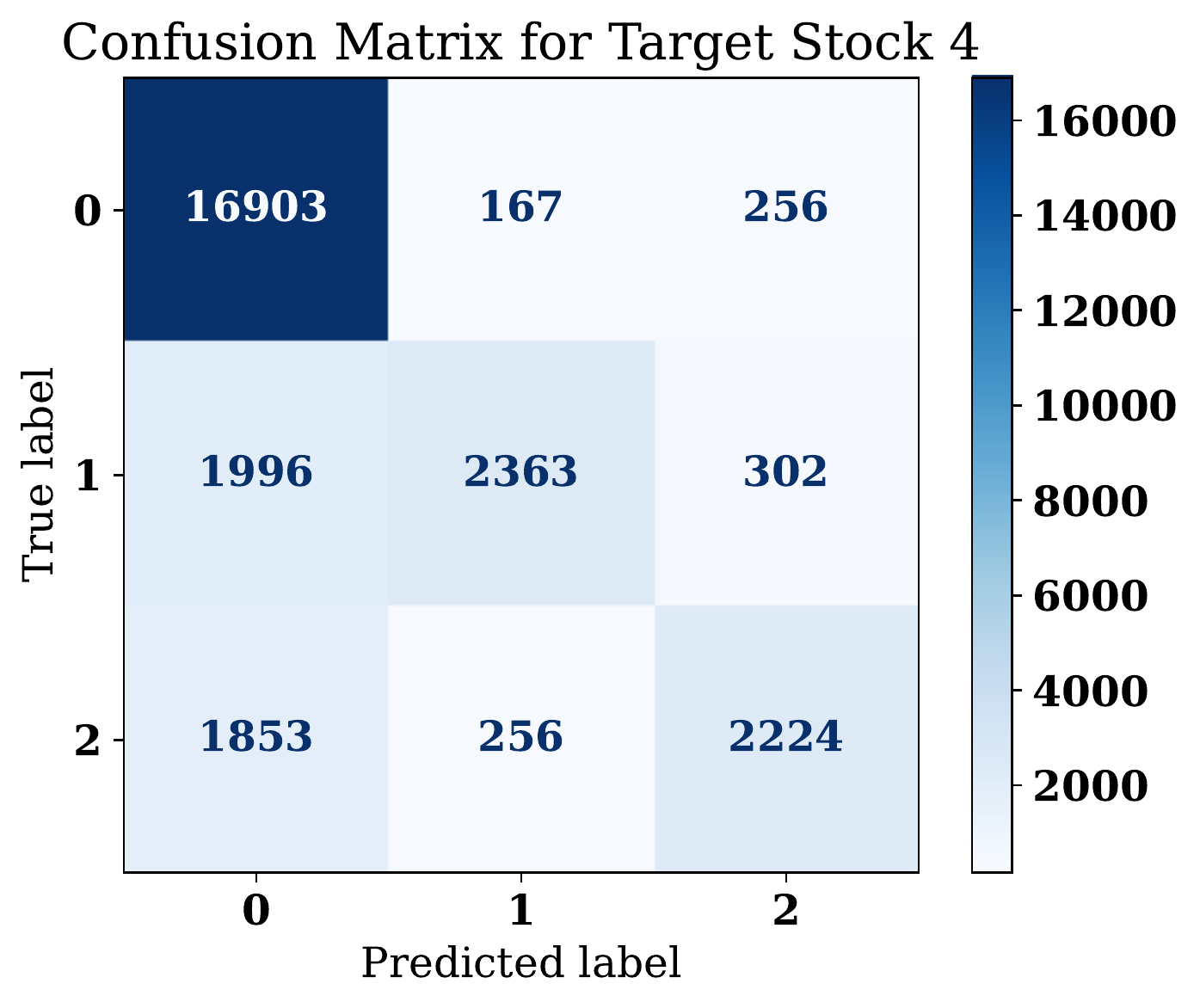}
    \end{minipage}
    \begin{minipage}[t]{.32\textwidth}
        \centering
        \includegraphics[width=1\textwidth]{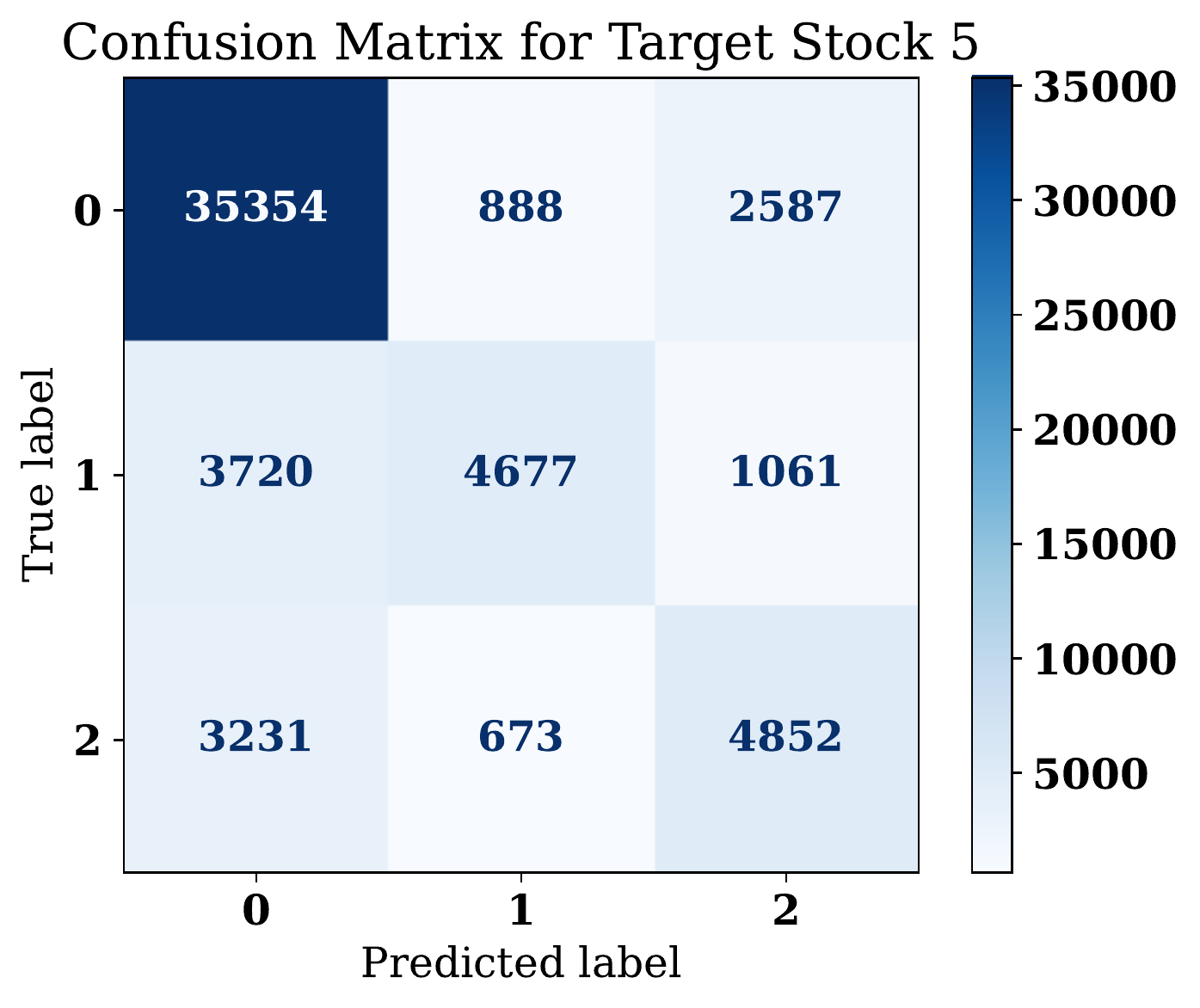}
        
    \end{minipage}
    
    \caption{Confusion matrices for the best model on each stock among the models in Tables \ref{table:experiments_Result} and \ref{table:experiments_Result_FCN}. Class 0 refers to the stock price remains the same, class 1 to the situation where the price goes up, and class 2 where the price goes downs.}
    \label{fig:CM_all_stocks}
\end{figure}

To have a better understanding of the prediction performance on each stock, we provide in Figure~\ref{fig:CM_all_stocks} the confusion matrices obtained by using the best model on each stock from the models listed in  Tables~\ref{table:experiments_Result} and \ref{table:experiments_Result_FCN}. As can be seen, for most of the stocks, the stationary class (label $0$) is the most populated one. The results show that for all stocks the majority of the predictions are to the stationary class (label $0$), while the models can distinguish between the stationary class and the two other classes (up and down corresponding to labels $1$ and $2$, respectively). Focusing on the classes up and down, we can see that when a time-series is classified to these two classes, it is correct in most of the cases, while misclassifications are mostly directed to the stationary class. When excluding the stationaly class, i.e., considering only the cases when the models predict that the mid-price will move up or down which are the cases that would lead to an action, distinguishing between class up (label $1$) and class down (label $2$) is very accurate. Another performance measure based on the provided confusion matrix is the win-rate, which is calculated by assuming that trades would only be placed when the forecast indicates that the stock price will change. As the prediction of changes is more difficult and important than the prediction of stationary class, the win-rate shows the performance of the proposed method in predicting class 1 and class 2. Table~\ref{table:win_rate_Result} shows the win-rate of all stocks' confusion matrices. As can be seen, our method achieves high performance in predicting the changes.

\begin{table*}[ht!]
\captionsetup{singlelinecheck = false, justification=justified}
\centering
\caption{Win-rate for each stock.}
\label{table:win_rate_Result}
\resizebox{0.95\linewidth}{!}{
\begin{tabular}{l|l|l|l|l|l|}
\cline{2-6}
                                   & Stock 1 & Stock 2 & Stock 3 & Stock 4 & Stock 5 \\ \hline
\multicolumn{1}{|l|}{Win-Rate(\%)} & 74.6    & 85.5    & 79.8    & 82.3    & 64.6    \\ \hline
\end{tabular}
}
\end{table*}

\begin{table*}[!ht]
\captionsetup{singlelinecheck = false, justification=justified}
\centering
\caption{Performance of TABL architectures (Mean $\pm$ Standard Deviation) measured on the test set in the Experiment Setup 2.}
\label{table:experiments_Result_setup_2}
\resizebox{0.95\linewidth}{!}{
\begin{tabular}{|l|c|c|c|c|c|}
\hline
\textbf{Method} &
  \textbf{Accuracy (\%)} &
  \textbf{Precision (\%)} &
  \textbf{Recall (\%)} &
  \textbf{F1-score (\%)} &
  \textbf{\#Params} \\ \hline
\texttt{TABL}        & 77.54 $\pm$ 01.20 & 69.26 $\pm$ 03.30 & 65.80 $\pm$ 01.10 & 67.33 $\pm$ 02.10 & 17,914\\ \hline
\texttt{TABL-fine-tune}    & 79.17 $\pm$ 01.30 & 72.56 $\pm$ 01.10 & 66.17 $\pm$ 03.10 & 68.77 $\pm$ 02.30 & 53,742 \\ \hline
\texttt{\textbf{aTABL-IS1}} & 
\textbf{80.31 $\pm$ 03.10} & 
\textbf{73.13 $\pm$ 02.40} & 
\textbf{66.26 $\pm$ 02.10} &
\textbf{69.73 $\pm$ 03.00} & 
\textbf{26,476} \\ \hline
\texttt{\textbf{aTABL-IS2}} &
  \textbf{80.56 $\pm$ 02.30} &
  \textbf{75.80 $\pm$ 01.10} &
  \textbf{66.47 $\pm$ 03.10} &
  \textbf{70.00 $\pm$ 03.20} &
  \textbf{25,636} \\ \hline
\end{tabular}}
\end{table*}

\begin{table*}[!ht]
\captionsetup{singlelinecheck = false, justification=justified}
\centering
\caption{Performance of CNN architectures (Mean $\pm$ Standard Deviation) measured on the test set in the Experiment Setup 2.}
\label{table:experiments_Result_setup_2_FCN}
\resizebox{0.95\linewidth}{!}{
\begin{tabular}{|l|l|l|l|l|c|}
\hline
\textbf{Method} & \multicolumn{1}{c|}{\textbf{Accuracy (\%)}} & \multicolumn{1}{c|}{\textbf{Precision (\%)}} & \multicolumn{1}{c|}{\textbf{Recall (\%)}} & \multicolumn{1}{c|}{\textbf{F1-score (\%)}} & \textbf{\#Params} \\ \hline
\texttt{CNN} & 72.33+-0.025 & 62.83+-0.024 & 62.43+-0.007 & 62.36+-0.015 & 17,091 \\ \hline
\texttt{\textbf{CNN-fine-tune}} & \textbf{73.89+-0.007} & \textbf{64.18+-0.01} & \textbf{63.71+-0.003} & \textbf{63.89+-0.005} & \textbf{51,273} \\ \hline
\texttt{aCNN} & 73.27+-0.01 & 63.62+-0.014 & 62.56+-0.005 & 62.94+-0.006 & 30,427  \\ \hline
\end{tabular}}
\end{table*}

\subsection{Experimental Setup 2}
To have a better real-world understanding of the advantages of the proposed method, we define another experimental setup as follows: the old dataset $\mathcal{D}_{\textrm{old}}$ consists of three stocks from the FI-2010 database and the new dataset $\mathcal{D}_{\textrm{new}}$ contains the remaining two stocks. We only have access to a pre-trained model $\mathcal{N}_{\textrm{old}}$ that has been trained on $\mathbf{T}_{\textrm{old}}$ but not to its training data $\mathcal{D}_{\textrm{old}}$. The objective is to build a model or a set of models that work well not only for the stocks in $\mathcal{D}_{\textrm{new}}$ but also for the stocks in $\mathcal{D}_{\textrm{old}}$. There are three approaches to tackle this problem: 
\begin{itemize}
    \item We simply keep the pre-trained model \texttt{TABL-base} or \texttt{CNN-base} and use it for all five stocks. The results related to this approach are denoted with \texttt{TABL-base} and \texttt{CNN-base}, respectively. 
    
    \item For each new stock in $\mathcal{D}_{\textrm{new}}$, we make a copy of the \texttt{TABL-base} or \texttt{CNN-base} model and fintune it using the data coming from the new stock. The fine-tuned model is used to generate predictions for the new stock. This approach requires the storage of three models: the pre-trained model trained on $\mathcal{D}_{\textrm{old}}$ and two models fine-tuned on data of the two new stocks $\mathcal{D}_{\textrm{new,4}}$ and $\mathcal{D}_{\textrm{new,5}}$. The results related to this approach are denoted with \texttt{TABL-fine-tune} and \texttt{CNN-fine-tune}. 
    
    \item Using the proposed method, in which for each of the new stocks an augmented model is created by adapting the pre-trained model using the corresponding datasets $\mathcal{D}_{\textrm{new,4}}$ and $\mathcal{D}_{\textrm{new,5}}$, respectively. Using our approach, we can store the pre-trained model $\mathcal{N}_{\textrm{old}}$ and only the auxiliary parameters for augmented models trained on $\mathcal{D}_{\textrm{new,4}}$ and $\mathcal{D}_{\textrm{new,5}}$. The results related to this approach are denoted with \texttt{aTABL-IS1} and  \texttt{aTABL-IS2}, corresponding to two implementations of augmented TABL, and the results for augmented CNNs are denoted with \texttt{aCNN}. 
\end{itemize}

With this experimental setup, we compare not only the prediction performance of three different approaches but also the storage cost associated with them. Tables~\ref{table:experiments_Result_setup_2} and~\ref{table:experiments_Result_setup_2_FCN} show the prediction performance as well as the total number of parameters that need to be stored for each case. From Table~\ref{table:experiments_Result_setup_2}, we can easily observe that the proposed method not only outperforms the finetuning approach in terms of performance but also in storage cost. In practice, when we have $N$ new stocks in our portfolio, the finetuning approach would require additional storage of $N$ models. On the other hand, for every new stock, our approach only requires a small fraction of additional storage for the auxiliary parameters. Even though the proposed augmentation has slightly inferior performance compared to the standard finetuning approach using CNNs, we still observe performance gains compared to the baseline CNN in Table \ref{table:experiments_Result_setup_2_FCN}. Regarding the storage cost, the proposed method always leads to storage savings compared to the finetuning approach.

\subsection{Online Learning Experimental Setup}\label{sec:FV_experiments}
Since the proposed model augmentation approach can also be used in an online learning setting in which new data of the same stock is generated through time, we also conducted experiments simulating this setting. More specifically, data from the first five days were used to train the base models. In order to make predictions for the eighth, ninth, and tenth days, there are three approaches:
\begin{itemize}
    \item We simply use the base models (\texttt{TABL-base} and \texttt{CNN-base}).
    \item We fine-tune the base models using data from the sixth and seventh days before making predictions for the last three days (results denoted with \texttt{TABL-fine-tune} and \texttt{CNN-fine-tune}).
    \item use our model augmentation method with the data from the sixth and seventh days to adapt the baseline models (results denoted with \texttt{aTABL-IS1}, \texttt{aTABL-IS2} and \texttt{aCNN}).
\end{itemize}

\begin{table*}[!ht]
\captionsetup{singlelinecheck = false, justification=justified}
\centering
\caption{Performance of CNN architectures (Mean $\pm$ Standard Deviation) measured on the test set in the online learning experimental setup}
\label{table:experiments_Result_setup_2_FCN_FV}
\resizebox{0.95\linewidth}{!}{
\begin{tabular}{|l|l|l|l|l|c|}
\hline
\textbf{Method} & \multicolumn{1}{c|}{\textbf{Accuracy (\%)}} & \multicolumn{1}{c|}{\textbf{Precision (\%)}} & \multicolumn{1}{c|}{\textbf{Recall (\%)}} & \multicolumn{1}{c|}{\textbf{F1-score (\%)}} & \textbf{Max rank} \\ \hline
\texttt{CNN-base} & 74.05$\pm$0.016 & 64.56$\pm$0.021 & 63.9$\pm$0.008 & 64.13$\pm$0.014 & - \\ \hline
\texttt{CNN-fine-tune} & 74.94$\pm$0.005 & 65.8$\pm$0.007 & 63.79$\pm$0.003 & 64.69$\pm$0.002 & - \\ \hline
\texttt{\textbf{aCNN}} & \textbf{75.24$\pm$0.013} & \textbf{66.26$\pm$0.02} & \textbf{63.94$\pm$0.003} & \textbf{64.93$\pm$0.007} & 1 \\ \hline
\end{tabular}}
\end{table*}

\begin{table*}[!ht]
\captionsetup{singlelinecheck = false, justification=justified}
\centering
\caption{Performance of TABL architectures (Mean $\pm$ Standard Deviation) measured on the test set in the online learning experimental setup}
\label{table:experiments_Result_setup_2_tabl_FV}
\resizebox{0.95\linewidth}{!}{
\begin{tabular}{|l|l|l|l|l|c|}
\hline
\textbf{Method} & \multicolumn{1}{c|}{\textbf{Accuracy (\%)}} & \multicolumn{1}{c|}{\textbf{Precision (\%)}} & \multicolumn{1}{c|}{\textbf{Recall (\%)}} & \multicolumn{1}{c|}{\textbf{F1-score (\%)}} & \textbf{Max rank} \\ \hline
\texttt{TABL-base} & 75.15$\pm$0.02 & 66.61$\pm$0.024 & 65.43$\pm$0.003 & 65.58$\pm$0.015 & - \\ \hline
\texttt{TABL-fine-tune} & 76.79$\pm$0.001 & 68.33$\pm$0.002 & 66.12$\pm$0.001 & 67.13$\pm$0.001 & - \\ \hline
\texttt{\textbf{aTABL-IS1}} & \textbf{79.88$\pm$0.02} & \textbf{76.16$\pm$0.045} & \textbf{64.93$\pm$0.006} & \textbf{68.68$\pm$0.018} & 9 \\ \hline
\texttt{\textbf{aTABL-IS2}} & \textbf{80.99$\pm$0.006} & \textbf{78.15$\pm$0.024} & \textbf{65.61$\pm$0.005} & \textbf{69.92$\pm$0.004} & 8 \\ \hline
\end{tabular}}
\end{table*}

Results obtained by following these three approaches are shown in Tables~\ref{table:experiments_Result_setup_2_FCN_FV} and~\ref{table:experiments_Result_setup_2_tabl_FV} and indicate that the proposed model augmentation method is also effective in adapting a pre-trained model in an online learning manner, yielding better performance compared to the baseline models as well as the fine-tuned models.

Here we should note that, in general, deep learning models may achieve different performances when trained using different random parameter initializations and when tested on data with different characteristics. We can see that the standard deviations of the performance of the models in this experiment, as well as in all other experiments reported above, are small in value. This shows that the adopted models provide reliable performance in terms of the stochastic nature of the Backpropagation algorithm.

\subsection{Trading Simulation}
An efficient trading system requires complex trading decisions and strategies. Forecasting price movements is an important part in a trading system. However, a more important component in a profitable system is the order placing strategy, i.e., when to place an order, at what price and volume to place the order and so on. The design of an efficient order placing strategy is out of the scope of this paper. This paper focuses on improving the prediction performance of mid-price direction movements, which can help develop more accurate trading systems. 

\begin{figure}[!htb]
\captionsetup{singlelinecheck = false, justification=justified}
    \begin{minipage}[t]{.32\textwidth}
        \centering
        \includegraphics[width=1\textwidth]{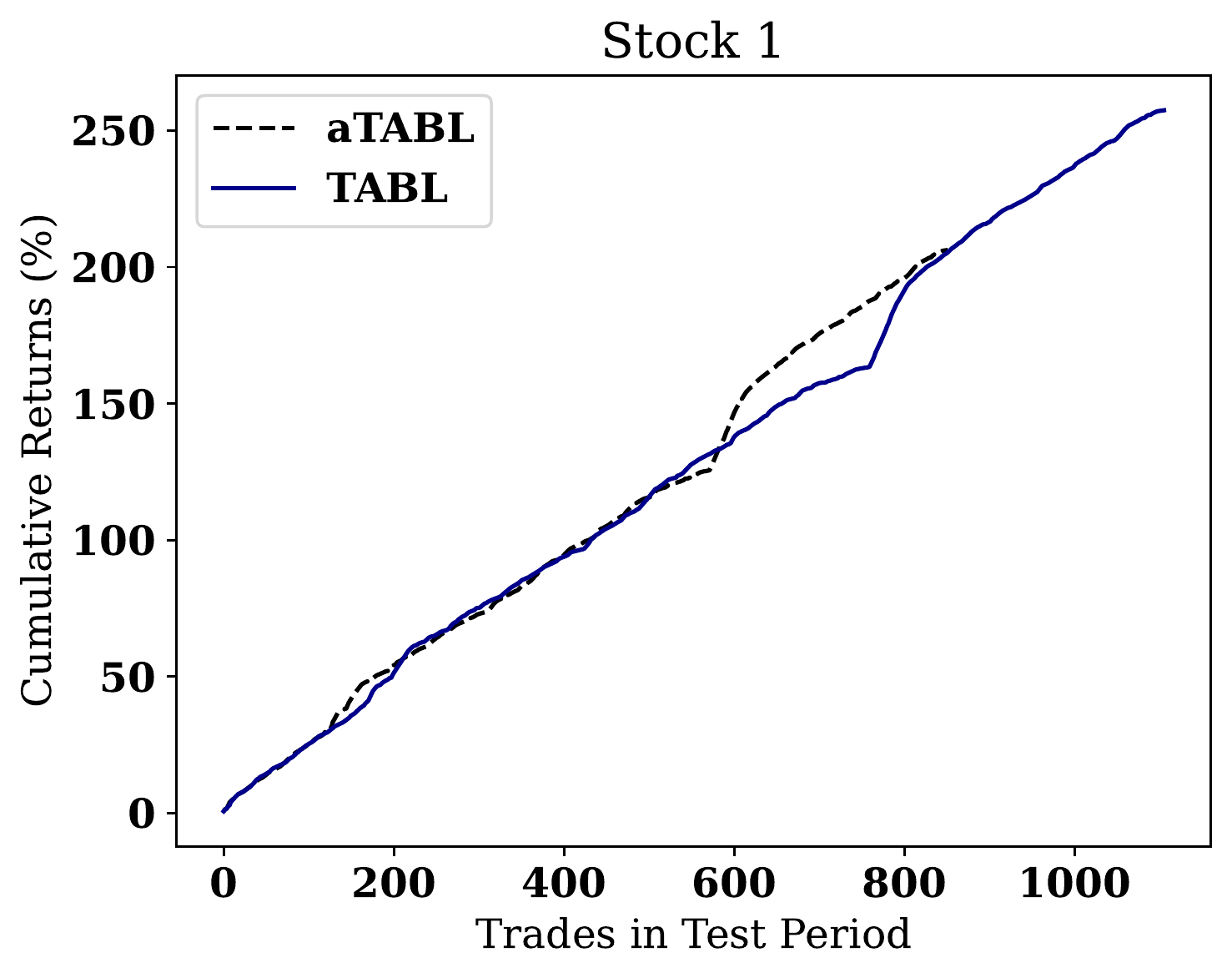}
    \end{minipage}
\hfill
    \begin{minipage}[t]{.32\textwidth}
        \centering
        \includegraphics[width=1\textwidth]{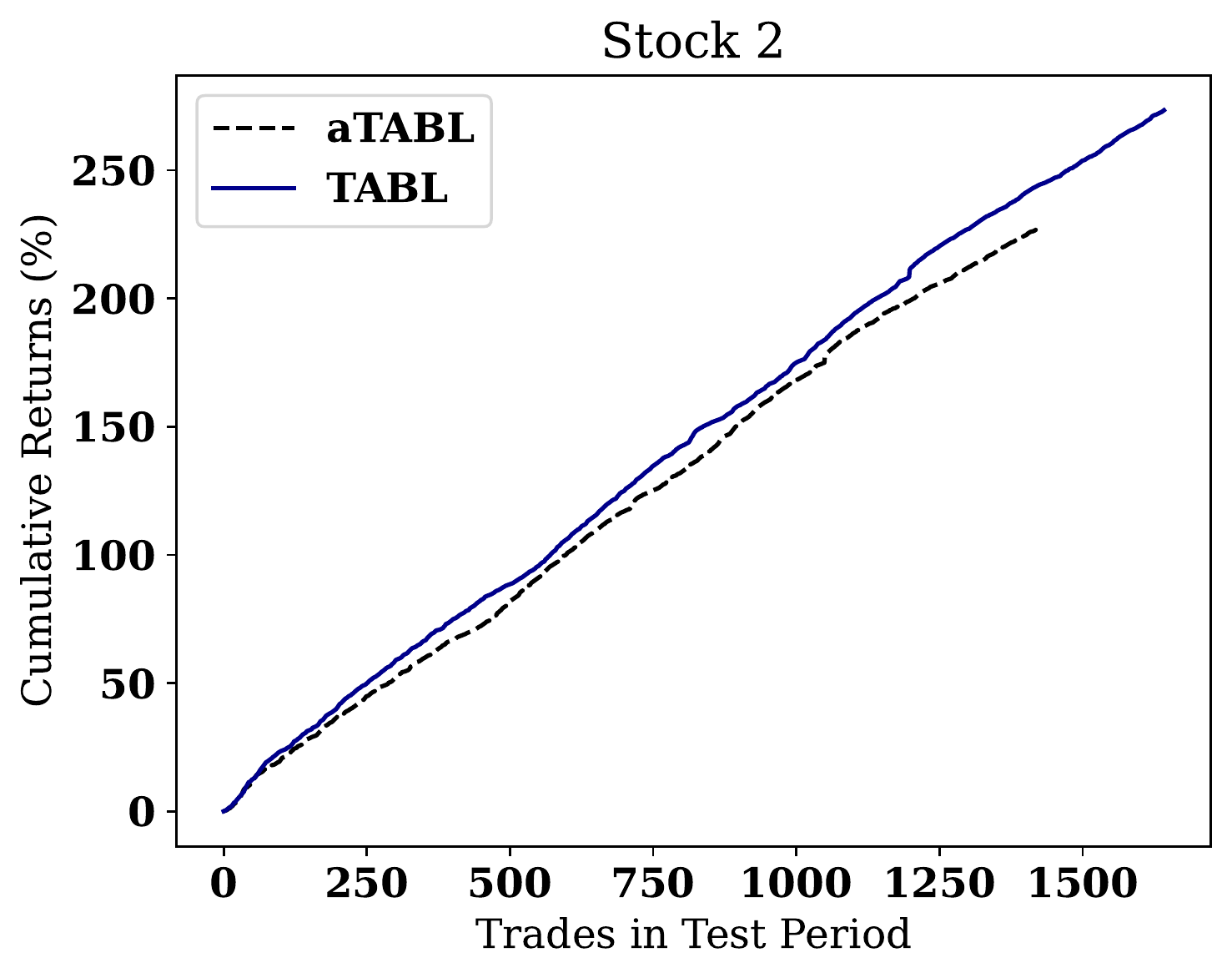}
        
    \end{minipage}
    \hfill
    \begin{minipage}[t]{.32\textwidth}
        \centering
        \includegraphics[width=1\textwidth]{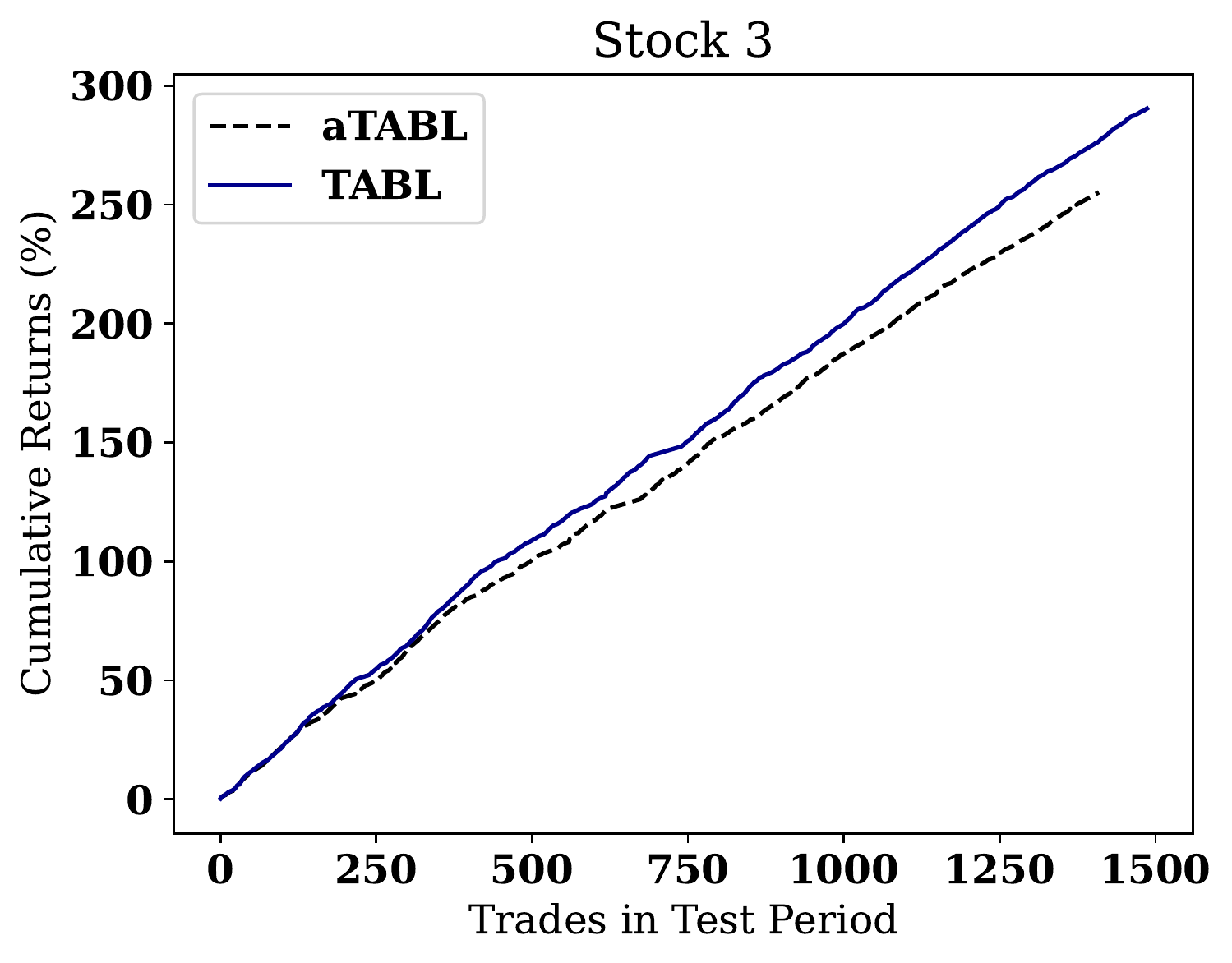}
        
    \end{minipage}
    \hfill
    \centering
    \begin{minipage}[t]{.32\textwidth}
        \centering
        \includegraphics[width=1\textwidth]{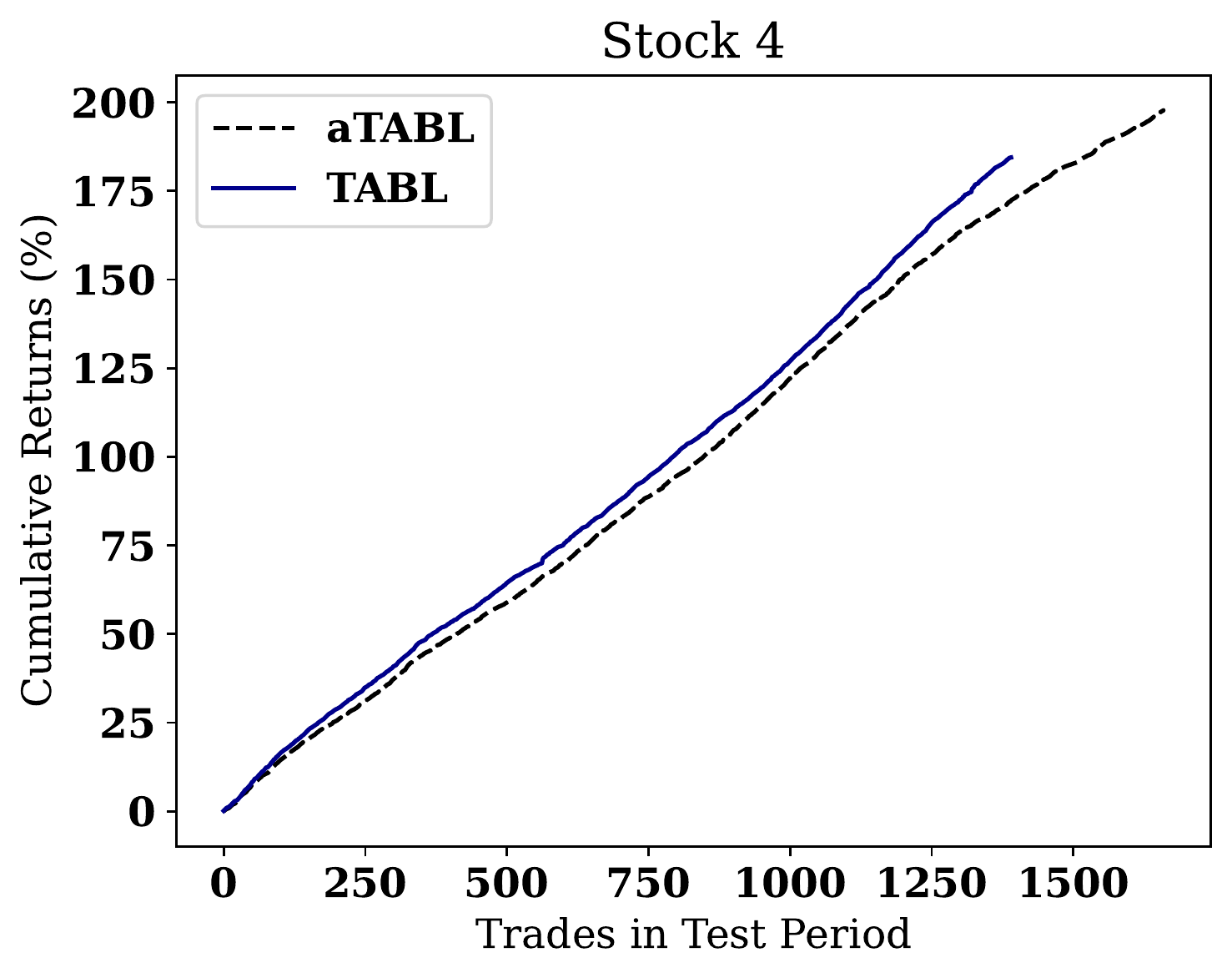}
    \end{minipage}
    \begin{minipage}[t]{.32\textwidth}
        \centering
        \includegraphics[width=1\textwidth]{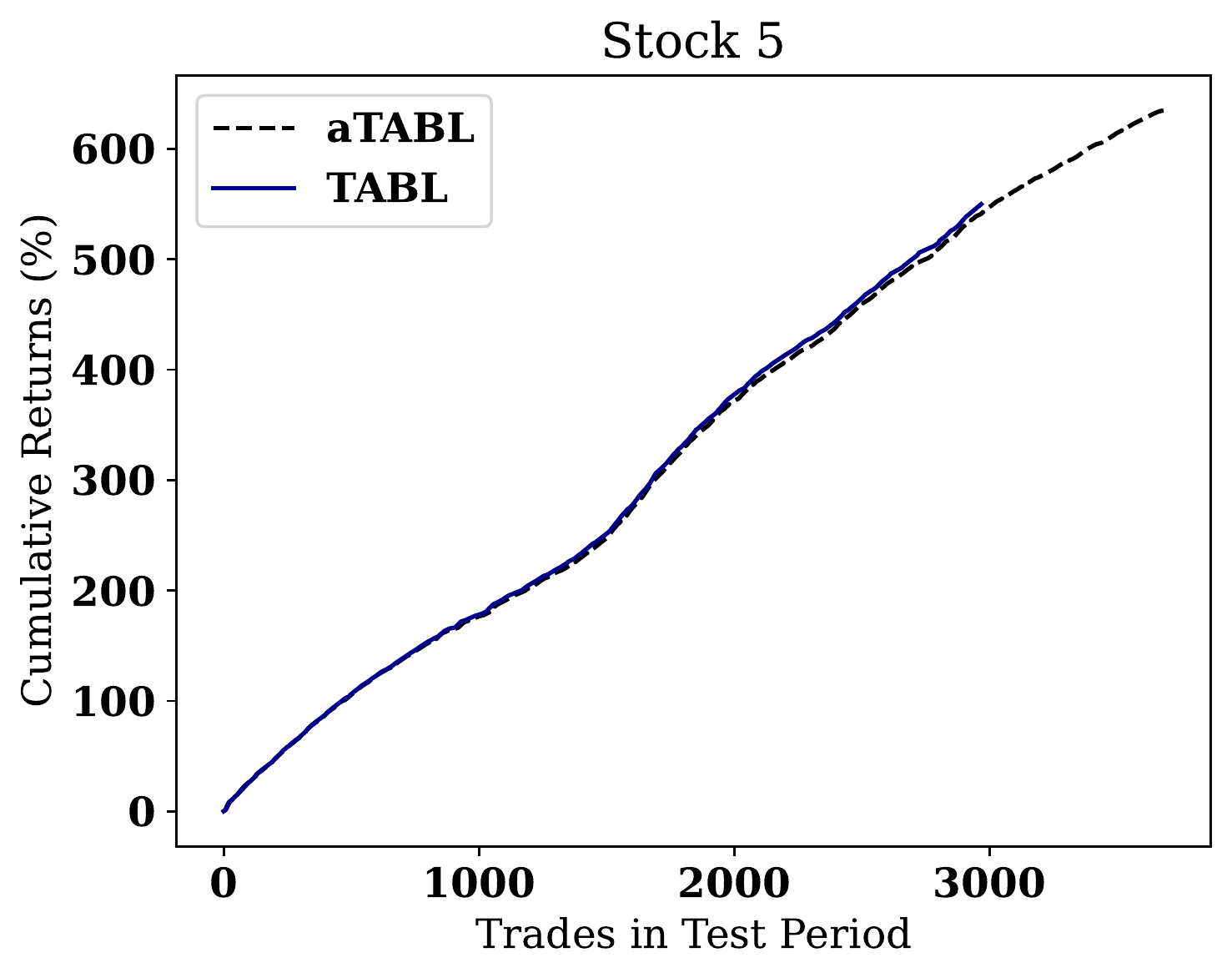}
        
    \end{minipage}
    
    \caption{Cumulative Returns (\%) of Trades in Test Period.}
    \label{fig:profit_all_stocks}
\end{figure}

To evaluate the profitability of the proposed model, we defined a simple long-only trading system with a naive order placing rule as follows: when the prediction is ``up" we will buy one share at best ask price and hold until the predicted label changes to ``down" (we do nothing for stationary predictions). At this point, we will sell at the best bid price. To calculate the actual cumulative return, we use the un-normalized ask and bid price values which are not included in the public FI-2010 dataset \citep{Ntakaris2017}. We do this because the normalization of the public data at the element-level leads to prices which can have negative values. The currency in the dataset is Euro and the transaction cost is not considered in this trading system. The plots in the Figure~\ref{fig:profit_all_stocks} show the cumulative returns of trades for the test dataset of each stock for the TABL and the best aTABL models in Table~\ref{table:experiments_Result}. The return of each trade is calculated as follows:
\begin{equation}
Returns  = (Exit_{Price}/Entry_{Price}) -1,
\end{equation}
where the $Exit_{Price}$ is the best bid price and $Entry_{Price}$ is the best ask price of the above trading strategy. 
The results in Figure~\ref{fig:profit_all_stocks} show that trading based on the predictions of TABL and aTABL is profitable. The plots of stocks 4 and 5 shows that the predictions of aTABL lead to better cumulative returns.

\section{Conclusion}\label{sec:Conclusions}
In this paper, we studied a new research problem defined on financial time-series data, that of efficiently training new models for stock mid-price time-series classification by exploiting knowledge in existing deep learning models trained on time-series data of different stocks or time-series data of past periods. We proposed a new method that exploits model augmentation and low-rank matrix approximation to improve the prediction performance and reduce the storage cost. Our model augmentation approach takes advantage of the learned information from the knowledge encoded in the parameters of an existing (pre-trained) model and learns auxiliary connections that are added to the pre-trained model to adapt it for the new task. The low rank approximation of auxiliary parameters regularizes the learning process and reduces the storage cost of new models. Extensive experiments on stock mid-price direction prediction tasks demonstrated that the proposed method can lead to performance improvements as well as a reduction in storage requirements during deployment. Interesting future research directions include investigating the effectiveness of the proposed approach in time-series classification problems coming from different applications, as well as the study of designing deep learning models targeting applications involving other forms of input data, like images, videos, audio, under the restrictions indicated by the adopted problem formulation.

\section*{Acknowledgment}
The research received funding from the Independent Research Fund Denmark project DISPA (Project Number: 9041-00004).

\appendix
\section{Appendix}\label{sec:Appendix}
Let us denote by $N$ the number of samples in a mini-batch during stochastic optimization. Below, we provide the computational complexity estimate for two implementation strategies described in Section 3. 
\\

\noindent \textbf{Computational Complexity of Implementation Strategy 1}
\begin{itemize}
    \item Computing Eq. (\ref{eq:bTABLa}) requires $ND D'T$ operations.
    \item Computing Eq. (\ref{eq:bTABLb}) requires $ND'TT$ operations.
    \item Computing Eq. (\ref{eq:bTABLe}) requires $ND'TT' + 2ND'T'$ operations.
\end{itemize}

\noindent \textbf{Computational Complexity of Implementation Strategy 2}
\begin{itemize}
    \item Computing Eq. (\ref{eq:cTABLa}) requires $N D D' T + N D K  T + N  K  T  D'$ operations.
    \item Computing Eq. (\ref{eq:cTABLb}) requires $ND'TT + ND'TK + ND'KT$ operations.
    \item Computing Eq. (\ref{eq:cTABLe}) requires $ND'TT' + ND'TK + ND'KT' + 2ND'T'$ operations.
\end{itemize}

\section{Appendix}\label{sec:Appendix_2}
To find the best topology for $\mathcal{N}_{\textrm{old}}$ for each target stock we ran experiments for different topologies and hyperparameter values. Tables \ref{tab:ab_1} - \ref{tab:ab_5} show the results of the topology ablation study for $\mathcal{N}_{\textrm{old}}$ of each target stock.
\begin{table}[!h]
\caption{Ablation study of $\mathcal{N}_{\textrm{old}}$ for Stock 1} 
\label{tab:ab_1} 
\centering
\resizebox{0.5\textwidth}{!}{%
\begin{tabular}{|l|l|}
\hline
Model                                                 & F1-score \\ \hline
layers\_{[}{[}60, 5{]}, {[}200, 10{]}{]}              & 0.694022 \\ \hline
layers\_{[}{[}60, 10{]}, {[}120, 5{]}{]}              & 0.687618 \\ \hline
layers\_{[}{[}60, 5{]}, {[}120, 10{]}, {[}50, 5{]}{]} & 0.686768 \\ \hline
layers\_{[}{[}60, 10{]}, {[}200, 10{]}{]}             & 0.68425  \\ \hline
layers\_{[}{[}50, 10{]}, {[}250, 5{]}{]}              & 0.68172  \\ \hline
layers\_{[}{[}60, 10{]}, {[}200, 5{]}{]}              & 0.67418  \\ \hline
\end{tabular}%
}
\end{table}

\begin{table}[!h]
\caption{Ablation study of $\mathcal{N}_{\textrm{old}}$ for Stock 2} 
\label{tab:ab_2} 
\centering
\resizebox{0.5\textwidth}{!}{%
\begin{tabular}{|l|l|}
\hline
Model                                                 & F1-score \\ \hline
layers\_{[}{[}60, 10{]}, {[}120, 5{]}{]}              & 0.701169 \\ \hline
layers\_{[}{[}60, 10{]}, {[}200, 5{]}{]}              & 0.693212 \\ \hline
layers\_{[}{[}60, 5{]}, {[}120, 10{]}, {[}50, 5{]}{]} & 0.692069 \\ \hline
layers\_{[}{[}60, 5{]}, {[}200, 10{]}{]}              & 0.685344 \\ \hline
layers\_{[}{[}60, 10{]}, {[}200, 10{]}{]}             & 0.68498  \\ \hline
layers\_{[}{[}50, 10{]}, {[}250, 5{]}{]}              & 0.683752 \\ \hline
\end{tabular}%
}
\end{table}

\begin{table}[!h]
\caption{Ablation study of $\mathcal{N}_{\textrm{old}}$ for Stock 3} 
\label{tab:ab_3} 
\centering
\resizebox{0.5\textwidth}{!}{%
\begin{tabular}{|l|l|}
\hline
Model                                                 & F1-score \\ \hline
layers\_{[}{[}60, 5{]}, {[}120, 10{]}, {[}50, 5{]}{]} & 0.701476 \\ \hline
layers\_{[}{[}60, 10{]}, {[}120, 5{]}{]}              & 0.699135 \\ \hline
layers\_{[}{[}60, 5{]}, {[}200, 10{]}{]}              & 0.698294 \\ \hline
layers\_{[}{[}60, 10{]}, {[}200, 5{]}{]}              & 0.698241 \\ \hline
layers\_{[}{[}50, 10{]}, {[}250, 5{]}{]}              & 0.698197 \\ \hline
layers\_{[}{[}60, 10{]}, {[}200, 10{]}{]}             & 0.689825 \\ \hline
\end{tabular}%
}
\end{table}

\begin{table}[!h]
\caption{Ablation study of $\mathcal{N}_{\textrm{old}}$ for Stock 4} 
\label{tab:ab_4} 
\centering
\resizebox{0.5\textwidth}{!}{%
\begin{tabular}{|l|l|}
\hline
Model                                                 & F1-score \\ \hline
layers\_{[}{[}60, 10{]}, {[}200, 10{]}{]}             & 0.698838 \\ \hline
layers\_{[}{[}60, 5{]}, {[}120, 10{]}, {[}50, 5{]}{]} & 0.69745  \\ \hline
layers\_{[}{[}60, 10{]}, {[}120, 5{]}{]}              & 0.697304 \\ \hline
layers\_{[}{[}60, 5{]}, {[}200, 10{]}{]}              & 0.691858 \\ \hline
layers\_{[}{[}50, 10{]}, {[}250, 5{]}{]}              & 0.687543 \\ \hline
layers\_{[}{[}60, 10{]}, {[}200, 5{]}{]}              & 0.685277 \\ \hline
\end{tabular}%
}
\end{table}

\begin{table}[!h]
\caption{Ablation study of $\mathcal{N}_{\textrm{old}}$ for Stock 5} 
\label{tab:ab_5} 
\centering
\resizebox{0.5\textwidth}{!}{%
\begin{tabular}{|l|l|}
\hline
Model                                                 & F1-score \\ \hline
layers\_{[}{[}60, 10{]}, {[}200, 10{]}{]}             & 0.710281 \\ \hline
layers\_{[}{[}60, 5{]}, {[}120, 10{]}, {[}50, 5{]}{]} & 0.709831 \\ \hline
layers\_{[}{[}60, 10{]}, {[}120, 5{]}{]}              & 0.706329 \\ \hline
layers\_{[}{[}50, 10{]}, {[}250, 5{]}{]}              & 0.706135 \\ \hline
layers\_{[}{[}60, 10{]}, {[}200, 5{]}{]}              & 0.70475  \\ \hline
layers\_{[}{[}60, 5{]}, {[}200, 10{]}{]}              & 0.70228  \\ \hline
\end{tabular}%
}
\end{table}

\FloatBarrier
\bibliography{refs}

\begin{thebibliography}{68}
\expandafter\ifx\csname natexlab\endcsname\relax\def\natexlab#1{#1}\fi
\providecommand{\url}[1]{\texttt{#1}}
\providecommand{\href}[2]{#2}
\providecommand{\path}[1]{#1}
\providecommand{\DOIprefix}{doi:}
\providecommand{\ArXivprefix}{arXiv:}
\providecommand{\URLprefix}{URL: }
\providecommand{\Pubmedprefix}{pmid:}
\providecommand{\doi}[1]{\href{http://dx.doi.org/#1}{\path{#1}}}
\providecommand{\Pubmed}[1]{\href{pmid:#1}{\path{#1}}}
\providecommand{\bibinfo}[2]{#2}
\ifx\xfnm\relax \def\xfnm[#1]{\unskip,\space#1}\fi
\bibitem[{Berndt \& Clifford(1994)}]{berndt1994using}
\bibinfo{author}{Berndt, D.~J.}, \& \bibinfo{author}{Clifford, J.}
  (\bibinfo{year}{1994}).
\newblock \bibinfo{title}{Using dynamic time warping to find patterns in time
  series.}
\newblock In {\it \bibinfo{booktitle}{KDD workshop}\/} (pp.
  \bibinfo{pages}{359--370}).
\newblock volume~\bibinfo{volume}{10}.
\bibitem[{Bulat et~al.(2020)Bulat, Kossaifi, Tzimiropoulos \&
  Pantic}]{bulat2020incremental}
\bibinfo{author}{Bulat, A.}, \bibinfo{author}{Kossaifi, J.},
  \bibinfo{author}{Tzimiropoulos, G.}, \& \bibinfo{author}{Pantic, M.}
  (\bibinfo{year}{2020}).
\newblock \bibinfo{title}{Incremental multi-domain learning with network latent
  tensor factorization}.
\newblock In {\it \bibinfo{booktitle}{AAAI}\/} (pp.
  \bibinfo{pages}{10470--10477}).
\bibitem[{Caruana(1997)}]{caruana1997multitask}
\bibinfo{author}{Caruana, R.} (\bibinfo{year}{1997}).
\newblock \bibinfo{title}{Multitask learning}.
\newblock {\it \bibinfo{journal}{Machine learning}\/},  {\it
  \bibinfo{volume}{28}\/}, \bibinfo{pages}{41--75}.
\bibitem[{Cavalcante \& Oliveira(2015)}]{cavalcante2015approach}
\bibinfo{author}{Cavalcante, R.~C.}, \& \bibinfo{author}{Oliveira, A.~L.}
  (\bibinfo{year}{2015}).
\newblock \bibinfo{title}{An approach to handle concept drift in financial time
  series based on extreme learning machines and explicit drift detection}.
\newblock In {\it \bibinfo{booktitle}{international joint conference on neural
  networks}\/} (pp. \bibinfo{pages}{1--8}).
\bibitem[{Cont(2011)}]{Cont2011}
\bibinfo{author}{Cont, R.} (\bibinfo{year}{2011}).
\newblock \bibinfo{title}{Statistical modeling of high-frequency financial
  data}.
\newblock {\it \bibinfo{journal}{IEEE Signal Processing Magazine}\/},  {\it
  \bibinfo{volume}{28}\/}, \bibinfo{pages}{16--25}.
\bibitem[{Denton et~al.(2014)Denton, Zaremba, Bruna, LeCun \&
  Fergus}]{denton2014exploiting}
\bibinfo{author}{Denton, E.~L.}, \bibinfo{author}{Zaremba, W.},
  \bibinfo{author}{Bruna, J.}, \bibinfo{author}{LeCun, Y.}, \&
  \bibinfo{author}{Fergus, R.} (\bibinfo{year}{2014}).
\newblock \bibinfo{title}{Exploiting linear structure within convolutional
  networks for efficient evaluation}.
\newblock In {\it \bibinfo{booktitle}{Advances in Neural Information Processing
  Systems}\/} (pp. \bibinfo{pages}{1269--1277}).
\bibitem[{Ding \& Fu(2018)}]{ding2018deep}
\bibinfo{author}{Ding, Z.}, \& \bibinfo{author}{Fu, Y.} (\bibinfo{year}{2018}).
\newblock \bibinfo{title}{Deep transfer low-rank coding for cross-domain
  learning}.
\newblock {\it \bibinfo{journal}{IEEE Transactions on Neural Networks and
  Learning Systems}\/},  {\it \bibinfo{volume}{30}\/},
  \bibinfo{pages}{1768--1779}.
\bibitem[{Ding et~al.(2016)Ding, Shao \& Fu}]{ding2016incomplete}
\bibinfo{author}{Ding, Z.}, \bibinfo{author}{Shao, M.}, \& \bibinfo{author}{Fu,
  Y.} (\bibinfo{year}{2016}).
\newblock \bibinfo{title}{Incomplete multisource transfer learning}.
\newblock {\it \bibinfo{journal}{IEEE Transactions on Neural Networks and
  Learning Systems}\/},  {\it \bibinfo{volume}{29}\/},
  \bibinfo{pages}{310--323}.
\bibitem[{Dixon(2017)}]{Dixon2017sequence}
\bibinfo{author}{Dixon, M.~F.} (\bibinfo{year}{2017}).
\newblock \bibinfo{title}{Sequence classification of the limit order book using
  recurrent neural networks}.
\newblock {\it \bibinfo{journal}{{SSRN} Electronic Journal}\/}, .
\bibitem[{Dredze et~al.(2010)Dredze, Kulesza \& Crammer}]{dredze2010multi}
\bibinfo{author}{Dredze, M.}, \bibinfo{author}{Kulesza, A.}, \&
  \bibinfo{author}{Crammer, K.} (\bibinfo{year}{2010}).
\newblock \bibinfo{title}{Multi-domain learning by confidence-weighted
  parameter combination}.
\newblock {\it \bibinfo{journal}{Machine Learning}\/},  {\it
  \bibinfo{volume}{79}\/}, \bibinfo{pages}{123--149}.
\bibitem[{Duan et~al.(2012)Duan, Xu \& Tsang}]{duan2012domain}
\bibinfo{author}{Duan, L.}, \bibinfo{author}{Xu, D.}, \&
  \bibinfo{author}{Tsang, I. W.-H.} (\bibinfo{year}{2012}).
\newblock \bibinfo{title}{Domain adaptation from multiple sources: A
  domain-dependent regularization approach}.
\newblock {\it \bibinfo{journal}{IEEE Transactions on Neural Networks and
  Learning Systems}\/},  {\it \bibinfo{volume}{23}\/},
  \bibinfo{pages}{504--518}.
\bibitem[{Fawaz et~al.(2018)Fawaz, Forestier, Weber, Idoumghar \&
  Muller}]{fawaz2018transfer}
\bibinfo{author}{Fawaz, H.~I.}, \bibinfo{author}{Forestier, G.},
  \bibinfo{author}{Weber, J.}, \bibinfo{author}{Idoumghar, L.}, \&
  \bibinfo{author}{Muller, P.-A.} (\bibinfo{year}{2018}).
\newblock \bibinfo{title}{Transfer learning for time series classification}.
\newblock In {\it \bibinfo{booktitle}{IEEE International Conference on Big
  Data}\/} (pp. \bibinfo{pages}{1367--1376}).
\bibitem[{Fons et~al.(2020)Fons, Dawson, Zeng, Keane \&
  Iosifidis}]{fons2020augmenting}
\bibinfo{author}{Fons, E.}, \bibinfo{author}{Dawson, P.},
  \bibinfo{author}{Zeng, X.-j.}, \bibinfo{author}{Keane, J.}, \&
  \bibinfo{author}{Iosifidis, A.} (\bibinfo{year}{2020}).
\newblock \bibinfo{title}{Augmenting transferred representations for stock
  classification}.
\newblock {\it \bibinfo{journal}{arXiv preprint arXiv:2011.04545}\/}, .
\bibitem[{Gama et~al.(2014)Gama, {\v{Z}}liobait{\.e}, Bifet, Pechenizkiy \&
  Bouchachia}]{gama2014survey}
\bibinfo{author}{Gama, J.}, \bibinfo{author}{{\v{Z}}liobait{\.e}, I.},
  \bibinfo{author}{Bifet, A.}, \bibinfo{author}{Pechenizkiy, M.}, \&
  \bibinfo{author}{Bouchachia, A.} (\bibinfo{year}{2014}).
\newblock \bibinfo{title}{A survey on concept drift adaptation}.
\newblock {\it \bibinfo{journal}{ACM Computing Surveys}\/},  {\it
  \bibinfo{volume}{46}\/}, \bibinfo{pages}{1--37}.
\bibitem[{Ganin et~al.(2016)Ganin, Ustinova, Ajakan, Germain, Larochelle,
  Laviolette, Marchand \& Lempitsky}]{ganin2016domain}
\bibinfo{author}{Ganin, Y.}, \bibinfo{author}{Ustinova, E.},
  \bibinfo{author}{Ajakan, H.}, \bibinfo{author}{Germain, P.},
  \bibinfo{author}{Larochelle, H.}, \bibinfo{author}{Laviolette, F.},
  \bibinfo{author}{Marchand, M.}, \& \bibinfo{author}{Lempitsky, V.}
  (\bibinfo{year}{2016}).
\newblock \bibinfo{title}{Domain-adversarial training of neural networks}.
\newblock {\it \bibinfo{journal}{The Journal of Machine Learning Research}\/},
  {\it \bibinfo{volume}{17}\/}, \bibinfo{pages}{2096--2030}.
\bibitem[{Ge et~al.(2013)Ge, Gao \& Zhang}]{ge2013oms}
\bibinfo{author}{Ge, L.}, \bibinfo{author}{Gao, J.}, \& \bibinfo{author}{Zhang,
  A.} (\bibinfo{year}{2013}).
\newblock \bibinfo{title}{Oms-tl: A framework of online multiple source
  transfer learning}.
\newblock In {\it \bibinfo{booktitle}{ACM International Conference on
  Information \& Knowledge Management}\/} (pp. \bibinfo{pages}{2423--2428}).
\bibitem[{Gretton et~al.(2012)Gretton, Borgwardt, Rasch, Sch{\"o}lkopf \&
  Smola}]{gretton2012kernel}
\bibinfo{author}{Gretton, A.}, \bibinfo{author}{Borgwardt, K.~M.},
  \bibinfo{author}{Rasch, M.~J.}, \bibinfo{author}{Sch{\"o}lkopf, B.}, \&
  \bibinfo{author}{Smola, A.} (\bibinfo{year}{2012}).
\newblock \bibinfo{title}{A kernel two-sample test}.
\newblock {\it \bibinfo{journal}{The Journal of Machine Learning Research}\/},
  {\it \bibinfo{volume}{13}\/}, \bibinfo{pages}{723--773}.
\bibitem[{He et~al.(2016)He, Zhang, Ren \& Sun}]{He_2016_CVPR}
\bibinfo{author}{He, K.}, \bibinfo{author}{Zhang, X.}, \bibinfo{author}{Ren,
  S.}, \& \bibinfo{author}{Sun, J.} (\bibinfo{year}{2016}).
\newblock \bibinfo{title}{Deep residual learning for image recognition}.
\newblock In {\it \bibinfo{booktitle}{IEEE Conference on Computer Vision and
  Pattern Recognition}\/}.
\bibitem[{Hedegaard et~al.(2021)Hedegaard, Sheikh-Omar \&
  Iosifidis}]{hedegaard2021supervised}
\bibinfo{author}{Hedegaard, L.}, \bibinfo{author}{Sheikh-Omar, O.}, \&
  \bibinfo{author}{Iosifidis, A.} (\bibinfo{year}{2021}).
\newblock \bibinfo{title}{{S}upervised {D}omain {A}daptation; a {G}raph
  {E}mbedding perspective and a rectified experimental protocol}.
\newblock {\it \bibinfo{journal}{IEEE Transactions on Image Processing}\/},
  {\it \bibinfo{volume}{30}\/}, \bibinfo{pages}{8619--8631}.
\bibitem[{Heidari \& Iosifidis(2020)}]{heidari2020progressive}
\bibinfo{author}{Heidari, N.}, \& \bibinfo{author}{Iosifidis, A.}
  (\bibinfo{year}{2020}).
\newblock \bibinfo{title}{Progressive spatio-temporal graph convolutional
  network for skeleton-based human action recognition}.
\newblock {\it \bibinfo{journal}{IEEE International Conference on Acoustics,
  Speech and Signal Processing}\/}, .
\bibitem[{Huang et~al.(2017)Huang, Liu, Van Der~Maaten \&
  Weinberger}]{huang2017densely}
\bibinfo{author}{Huang, G.}, \bibinfo{author}{Liu, Z.}, \bibinfo{author}{Van
  Der~Maaten, L.}, \& \bibinfo{author}{Weinberger, K.~Q.}
  (\bibinfo{year}{2017}).
\newblock \bibinfo{title}{Densely connected convolutional networks}.
\newblock In {\it \bibinfo{booktitle}{IEEE Conference on Computer Vision and
  Pattern Recognition}\/} (pp. \bibinfo{pages}{4700--4708}).
\bibitem[{Huang \& Yu(2018)}]{huang2018ltnn}
\bibinfo{author}{Huang, H.}, \& \bibinfo{author}{Yu, H.}
  (\bibinfo{year}{2018}).
\newblock \bibinfo{title}{Ltnn: A layerwise tensorized compression of
  multilayer neural network}.
\newblock {\it \bibinfo{journal}{IEEE transactions on neural networks and
  learning systems}\/},  {\it \bibinfo{volume}{30}\/},
  \bibinfo{pages}{1497--1511}.
\bibitem[{Jaderberg et~al.(2014)Jaderberg, Vedaldi \&
  Zisserman}]{jaderberg2014speeding}
\bibinfo{author}{Jaderberg, M.}, \bibinfo{author}{Vedaldi, A.}, \&
  \bibinfo{author}{Zisserman, A.} (\bibinfo{year}{2014}).
\newblock \bibinfo{title}{Speeding up convolutional neural networks with low
  rank expansions}.
\newblock {\it \bibinfo{journal}{arXiv preprint arXiv:1405.3866}\/}, .
\bibitem[{Kanjamapornkul et~al.(2016)Kanjamapornkul, Pin{\v{c}}{\'a}k \&
  Barto{\v{s}}}]{kanjamapornkul2016study}
\bibinfo{author}{Kanjamapornkul, K.}, \bibinfo{author}{Pin{\v{c}}{\'a}k, R.},
  \& \bibinfo{author}{Barto{\v{s}}, E.} (\bibinfo{year}{2016}).
\newblock \bibinfo{title}{The study of thai stock market across the 2008
  financial crisis}.
\newblock {\it \bibinfo{journal}{Physica A: Statistical Mechanics and its
  Applications}\/},  {\it \bibinfo{volume}{462}\/}, \bibinfo{pages}{117--133}.
\bibitem[{Kanjamapornkul et~al.(2017)Kanjamapornkul, Pin{\v{c}}{\'a}k,
  Chunithipaisan \& Barto{\v{s}}}]{kanjamapornkul2017support}
\bibinfo{author}{Kanjamapornkul, K.}, \bibinfo{author}{Pin{\v{c}}{\'a}k, R.},
  \bibinfo{author}{Chunithipaisan, S.}, \& \bibinfo{author}{Barto{\v{s}}, E.}
  (\bibinfo{year}{2017}).
\newblock \bibinfo{title}{Support spinor machine}.
\newblock {\it \bibinfo{journal}{Digital Signal Processing}\/},  {\it
  \bibinfo{volume}{70}\/}, \bibinfo{pages}{59--72}.
\bibitem[{Kiranyaz et~al.(2017)Kiranyaz, Ince, Iosifidis \&
  Gabbouj}]{kiranyaz2017progressive}
\bibinfo{author}{Kiranyaz, S.}, \bibinfo{author}{Ince, T.},
  \bibinfo{author}{Iosifidis, A.}, \& \bibinfo{author}{Gabbouj, M.}
  (\bibinfo{year}{2017}).
\newblock \bibinfo{title}{Progressive operational perceptrons}.
\newblock {\it \bibinfo{journal}{Neurocomputing}\/},  {\it
  \bibinfo{volume}{224}\/}, \bibinfo{pages}{142--154}.
\bibitem[{Kiranyaz et~al.(2020)Kiranyaz, Ince, Iosifidis \&
  Gabbouj}]{kiranyaz2020operational}
\bibinfo{author}{Kiranyaz, S.}, \bibinfo{author}{Ince, T.},
  \bibinfo{author}{Iosifidis, A.}, \& \bibinfo{author}{Gabbouj, M.}
  (\bibinfo{year}{2020}).
\newblock \bibinfo{title}{Operational neural networks}.
\newblock {\it \bibinfo{journal}{Neural Computing and Applications}\/},  (pp.
  \bibinfo{pages}{1--24}).
\bibitem[{Kolda \& Bader(2009)}]{kolda2009tensor}
\bibinfo{author}{Kolda, T.~G.}, \& \bibinfo{author}{Bader, B.~W.}
  (\bibinfo{year}{2009}).
\newblock \bibinfo{title}{Tensor decompositions and applications}.
\newblock {\it \bibinfo{journal}{SIAM review}\/},  {\it
  \bibinfo{volume}{51}\/}, \bibinfo{pages}{455--500}.
\bibitem[{Koshiyama et~al.(2020)Koshiyama, Flennerhag, Blumberg, Firoozye \&
  Treleaven}]{koshiyama2020quantnet}
\bibinfo{author}{Koshiyama, A.}, \bibinfo{author}{Flennerhag, S.},
  \bibinfo{author}{Blumberg, S.~B.}, \bibinfo{author}{Firoozye, N.}, \&
  \bibinfo{author}{Treleaven, P.} (\bibinfo{year}{2020}).
\newblock \bibinfo{title}{Quantnet: Transferring learning across systematic
  trading strategies}.
\newblock {\it \bibinfo{journal}{arXiv preprint arXiv:2004.03445}\/}, .
\bibitem[{Long et~al.(2015)Long, Cao, Wang \& Jordan}]{long2015learning}
\bibinfo{author}{Long, M.}, \bibinfo{author}{Cao, Y.}, \bibinfo{author}{Wang,
  J.}, \& \bibinfo{author}{Jordan, M.} (\bibinfo{year}{2015}).
\newblock \bibinfo{title}{Learning transferable features with deep adaptation
  networks}.
\newblock In {\it \bibinfo{booktitle}{International Conference on Machine
  Learning}\/} (pp. \bibinfo{pages}{97--105}).
\bibitem[{Minh et~al.(2018)Minh, Sadeghi-Niaraki, Huy, Min \& Moon}]{Minh2018}
\bibinfo{author}{Minh, D.~L.}, \bibinfo{author}{Sadeghi-Niaraki, A.},
  \bibinfo{author}{Huy, H.~D.}, \bibinfo{author}{Min, K.}, \&
  \bibinfo{author}{Moon, H.} (\bibinfo{year}{2018}).
\newblock \bibinfo{title}{Deep learning approach for short-term stock trends
  prediction based on two-stream gated recurrent unit network}.
\newblock {\it \bibinfo{journal}{{IEEE} Access}\/},  {\it
  \bibinfo{volume}{6}\/}, \bibinfo{pages}{55392--55404}.
\bibitem[{Nguyen \& Yoon(2019)}]{nguyen2019novel}
\bibinfo{author}{Nguyen, T.-T.}, \& \bibinfo{author}{Yoon, S.}
  (\bibinfo{year}{2019}).
\newblock \bibinfo{title}{A novel approach to short-term stock price movement
  prediction using transfer learning}.
\newblock {\it \bibinfo{journal}{Applied Sciences}\/},  {\it
  \bibinfo{volume}{9}\/}, \bibinfo{pages}{4745}.
\bibitem[{Ntakaris et~al.(2018)Ntakaris, Magris, Kanniainen, Gabbouj \&
  Iosifidis}]{Ntakaris2017}
\bibinfo{author}{Ntakaris, A.}, \bibinfo{author}{Magris, M.},
  \bibinfo{author}{Kanniainen, J.}, \bibinfo{author}{Gabbouj, M.}, \&
  \bibinfo{author}{Iosifidis, A.} (\bibinfo{year}{2018}).
\newblock \bibinfo{title}{Benchmark dataset for mid-price forecasting of limit
  order book data with machine learning methods}.
\newblock {\it \bibinfo{journal}{Journal of Forecasting}\/},  {\it
  \bibinfo{volume}{37}\/}, \bibinfo{pages}{852--866}.
\bibitem[{Pan \& Yang(2009)}]{pan2009survey}
\bibinfo{author}{Pan, S.~J.}, \& \bibinfo{author}{Yang, Q.}
  (\bibinfo{year}{2009}).
\newblock \bibinfo{title}{A survey on transfer learning}.
\newblock {\it \bibinfo{journal}{IEEE Transactions on Knowledge and Data
  Engineering}\/},  {\it \bibinfo{volume}{22}\/}, \bibinfo{pages}{1345--1359}.
\bibitem[{{Passalis} et~al.(2020){Passalis}, {Tefas}, {Kanniainen}, {Gabbouj}
  \& {Iosifidis}}]{passalis2019deep}
\bibinfo{author}{{Passalis}, N.}, \bibinfo{author}{{Tefas}, A.},
  \bibinfo{author}{{Kanniainen}, J.}, \bibinfo{author}{{Gabbouj}, M.}, \&
  \bibinfo{author}{{Iosifidis}, A.} (\bibinfo{year}{2020}).
\newblock \bibinfo{title}{Deep adaptive input normalization for time series
  forecasting}.
\newblock {\it \bibinfo{journal}{IEEE Transactions on Neural Networks and
  Learning Systems}\/},  {\it \bibinfo{volume}{31}\/},
  \bibinfo{pages}{3760--3765}.
\bibitem[{Passalis et~al.(2020)Passalis, Tefas, Kanniainen, Gabbouj \&
  Iosifidis}]{Passalis2020}
\bibinfo{author}{Passalis, N.}, \bibinfo{author}{Tefas, A.},
  \bibinfo{author}{Kanniainen, J.}, \bibinfo{author}{Gabbouj, M.}, \&
  \bibinfo{author}{Iosifidis, A.} (\bibinfo{year}{2020}).
\newblock \bibinfo{title}{Temporal logistic neural bag-of-features for
  financial time series forecasting leveraging limit order book data}.
\newblock {\it \bibinfo{journal}{Pattern Recognition Letters}\/},  {\it
  \bibinfo{volume}{136}\/}, \bibinfo{pages}{183--189}.
\bibitem[{Pratama et~al.(2019{\natexlab{a}})Pratama, de~Carvalho, Xie, Lughofer
  \& Lu}]{pratama2019atl}
\bibinfo{author}{Pratama, M.}, \bibinfo{author}{de~Carvalho, M.},
  \bibinfo{author}{Xie, R.}, \bibinfo{author}{Lughofer, E.}, \&
  \bibinfo{author}{Lu, J.} (\bibinfo{year}{2019}{\natexlab{a}}).
\newblock \bibinfo{title}{Atl: Autonomous knowledge transfer from many
  streaming processes}.
\newblock In {\it \bibinfo{booktitle}{ACM International Conference on
  Information and Knowledge Management}\/} (pp. \bibinfo{pages}{269--278}).
\bibitem[{Pratama et~al.(2019{\natexlab{b}})Pratama, de~Carvalho, Xie, Lughofer
  \& Lu}]{mahardhika2019autonomous}
\bibinfo{author}{Pratama, M.}, \bibinfo{author}{de~Carvalho, M.},
  \bibinfo{author}{Xie, R.}, \bibinfo{author}{Lughofer, E.}, \&
  \bibinfo{author}{Lu, J.} (\bibinfo{year}{2019}{\natexlab{b}}).
\newblock \bibinfo{title}{Atl: Autonomous knowledge transfer from many
  streaming processes}.
\newblock {\it \bibinfo{journal}{ACM International Conference on Information
  and Knowledge Management}\/},  (pp. \bibinfo{pages}{269--278}).
\bibitem[{Pratama et~al.(2019{\natexlab{c}})Pratama, Za'in, Ashfahani, Ong \&
  Ding}]{pratama2019automatic}
\bibinfo{author}{Pratama, M.}, \bibinfo{author}{Za'in, C.},
  \bibinfo{author}{Ashfahani, A.}, \bibinfo{author}{Ong, Y.~S.}, \&
  \bibinfo{author}{Ding, W.} (\bibinfo{year}{2019}{\natexlab{c}}).
\newblock \bibinfo{title}{Automatic construction of multi-layer perceptron
  network from streaming examples}.
\newblock In {\it \bibinfo{booktitle}{ACM International Conference on
  Information and Knowledge Management}\/} (pp. \bibinfo{pages}{1171--1180}).
\bibitem[{Renchunzi \& Pratama(2022)}]{renchunzi2022automatic}
\bibinfo{author}{Renchunzi, X.}, \& \bibinfo{author}{Pratama, M.}
  (\bibinfo{year}{2022}).
\newblock \bibinfo{title}{Automatic online multi-source domain adaptation}.
\newblock {\it \bibinfo{journal}{Information Sciences}\/},  {\it
  \bibinfo{volume}{582}\/}, \bibinfo{pages}{480--494}.
\bibitem[{Rosenstein et~al.(2005)Rosenstein, Marx, Kaelbling \&
  Dietterich}]{rosenstein2005transfer}
\bibinfo{author}{Rosenstein, M.~T.}, \bibinfo{author}{Marx, Z.},
  \bibinfo{author}{Kaelbling, L.~P.}, \& \bibinfo{author}{Dietterich, T.~G.}
  (\bibinfo{year}{2005}).
\newblock \bibinfo{title}{To transfer or not to transfer}.
\newblock In {\it \bibinfo{booktitle}{NIPS 2005 Workshop on Transfer
  Learning}\/} (pp. \bibinfo{pages}{1--4}).
\newblock volume \bibinfo{volume}{898}.
\bibitem[{Ross et~al.(2008)Ross, Lim, Lin \& Yang}]{ross2008incremental}
\bibinfo{author}{Ross, D.~A.}, \bibinfo{author}{Lim, J.}, \bibinfo{author}{Lin,
  R.-S.}, \& \bibinfo{author}{Yang, M.-H.} (\bibinfo{year}{2008}).
\newblock \bibinfo{title}{Incremental learning for robust visual tracking}.
\newblock {\it \bibinfo{journal}{International Journal of Computer Vision}\/},
  {\it \bibinfo{volume}{77}\/}, \bibinfo{pages}{125--141}.
\bibitem[{Ruan et~al.(2020)Ruan, Liu, Yuan, Li, Hu, Li \&
  Maybank}]{ruan2020edp}
\bibinfo{author}{Ruan, X.}, \bibinfo{author}{Liu, Y.}, \bibinfo{author}{Yuan,
  C.}, \bibinfo{author}{Li, B.}, \bibinfo{author}{Hu, W.}, \bibinfo{author}{Li,
  Y.}, \& \bibinfo{author}{Maybank, S.} (\bibinfo{year}{2020}).
\newblock \bibinfo{title}{Edp: An efficient decomposition and pruning scheme
  for convolutional neural network compression}.
\newblock {\it \bibinfo{journal}{IEEE Transactions on Neural Networks and
  Learning Systems}\/},  {\it \bibinfo{volume}{32}\/},
  \bibinfo{pages}{4499--4513}.
\bibitem[{Senhaji et~al.(2020)Senhaji, Raitoharju, Gabbouj \&
  Iosifidis}]{senhaji2020not}
\bibinfo{author}{Senhaji, A.}, \bibinfo{author}{Raitoharju, J.},
  \bibinfo{author}{Gabbouj, M.}, \& \bibinfo{author}{Iosifidis, A.}
  (\bibinfo{year}{2020}).
\newblock \bibinfo{title}{Not all domains are equally complex: Adaptive
  multi-domain learning}.
\newblock {\it \bibinfo{journal}{Internatonal Conference on Pattern
  Recognition}\/}, .
\bibitem[{Sezer et~al.(2020)Sezer, Gudelek \& Ozbayoglu}]{sezer2020financial}
\bibinfo{author}{Sezer, O.~B.}, \bibinfo{author}{Gudelek, M.~U.}, \&
  \bibinfo{author}{Ozbayoglu, A.~M.} (\bibinfo{year}{2020}).
\newblock \bibinfo{title}{Financial time series forecasting with deep learning:
  A systematic literature review: 2005--2019}.
\newblock {\it \bibinfo{journal}{Applied Soft Computing}\/},  {\it
  \bibinfo{volume}{90}\/}, \bibinfo{pages}{106181}.
\bibitem[{Shao et~al.(2014)Shao, Zhu \& Li}]{shao2014transfer}
\bibinfo{author}{Shao, L.}, \bibinfo{author}{Zhu, F.}, \& \bibinfo{author}{Li,
  X.} (\bibinfo{year}{2014}).
\newblock \bibinfo{title}{Transfer learning for visual categorization: A
  survey}.
\newblock {\it \bibinfo{journal}{IEEE Transactions on Neural Networks and
  Learning Systems}\/},  {\it \bibinfo{volume}{26}\/},
  \bibinfo{pages}{1019--1034}.
\bibitem[{Sirignano(2019)}]{sirignano2019deep}
\bibinfo{author}{Sirignano, J.~A.} (\bibinfo{year}{2019}).
\newblock \bibinfo{title}{Deep learning for limit order books}.
\newblock {\it \bibinfo{journal}{Quantitative Finance}\/},  {\it
  \bibinfo{volume}{19}\/}, \bibinfo{pages}{549--570}.
\bibitem[{Tran \& Iosifidis(2019)}]{tran2019learning}
\bibinfo{author}{Tran, D.~T.}, \& \bibinfo{author}{Iosifidis, A.}
  (\bibinfo{year}{2019}).
\newblock \bibinfo{title}{Learning to rank: A progressive neural network
  learning approach}.
\newblock In {\it \bibinfo{booktitle}{IEEE International Conference on
  Acoustics, Speech and Signal Processing}\/} (pp.
  \bibinfo{pages}{8355--8359}).
\bibitem[{Tran et~al.(2018)Tran, Iosifidis \& Gabbouj}]{tran2018improving}
\bibinfo{author}{Tran, D.~T.}, \bibinfo{author}{Iosifidis, A.}, \&
  \bibinfo{author}{Gabbouj, M.} (\bibinfo{year}{2018}).
\newblock \bibinfo{title}{Improving efficiency in convolutional neural networks
  with multilinear filters}.
\newblock {\it \bibinfo{journal}{Neural Networks}\/},  {\it
  \bibinfo{volume}{105}\/}, \bibinfo{pages}{328--339}.
\bibitem[{Tran et~al.(2019{\natexlab{a}})Tran, Iosifidis, Kanniainen \&
  Gabbouj}]{Tran2019a}
\bibinfo{author}{Tran, D.~T.}, \bibinfo{author}{Iosifidis, A.},
  \bibinfo{author}{Kanniainen, J.}, \& \bibinfo{author}{Gabbouj, M.}
  (\bibinfo{year}{2019}{\natexlab{a}}).
\newblock \bibinfo{title}{Temporal attention-augmented bilinear network for
  financial time-series data analysis}.
\newblock {\it \bibinfo{journal}{IEEE Transactions on Neural Networks and
  Learning Systems}\/},  {\it \bibinfo{volume}{30}\/},
  \bibinfo{pages}{1407--1418}.
\bibitem[{Tran et~al.(2019{\natexlab{b}})Tran, Kanniainen, Gabbouj \&
  Iosifidis}]{Tran2019}
\bibinfo{author}{Tran, D.~T.}, \bibinfo{author}{Kanniainen, J.},
  \bibinfo{author}{Gabbouj, M.}, \& \bibinfo{author}{Iosifidis, A.}
  (\bibinfo{year}{2019}{\natexlab{b}}).
\newblock \bibinfo{title}{Data-driven neural architecture learning for
  financial time-series forecasting}.
\newblock {\it \bibinfo{journal}{ArXiv}\/},  {\it
  \bibinfo{volume}{abs/1903.06751}\/}.
\bibitem[{{Tran} et~al.(2021){Tran}, {Kanniainen}, {Gabbouj} \&
  {Iosifidis}}]{tran2021bilinearInputNormalization}
\bibinfo{author}{{Tran}, D.~T.}, \bibinfo{author}{{Kanniainen}, J.},
  \bibinfo{author}{{Gabbouj}, M.}, \& \bibinfo{author}{{Iosifidis}, A.}
  (\bibinfo{year}{2021}).
\newblock \bibinfo{title}{Bilinear input normalization for neural networks in
  financial forecasting}.
\newblock {\it \bibinfo{journal}{arXiv:2109.00983}\/}, .
\bibitem[{Tran et~al.(2021)Tran, Kanniainen \& Iosifidis}]{tran2021informative}
\bibinfo{author}{Tran, D.~T.}, \bibinfo{author}{Kanniainen, J.}, \&
  \bibinfo{author}{Iosifidis, A.} (\bibinfo{year}{2021}).
\newblock \bibinfo{title}{How informative is the order book beyond the best
  levels? machine learning perspective}.
\newblock {\it \bibinfo{journal}{NeurIPS 2021 Workshop on Machine Learning
  meets Econometrics}\/}, .
\bibitem[{Tran et~al.(2019{\natexlab{c}})Tran, Kiranyaz, Gabbouj \&
  Iosifidis}]{tran2019heterogeneous}
\bibinfo{author}{Tran, D.~T.}, \bibinfo{author}{Kiranyaz, S.},
  \bibinfo{author}{Gabbouj, M.}, \& \bibinfo{author}{Iosifidis, A.}
  (\bibinfo{year}{2019}{\natexlab{c}}).
\newblock \bibinfo{title}{Heterogeneous multilayer generalized operational
  perceptron}.
\newblock {\it \bibinfo{journal}{IEEE Transactions on Neural Networks and
  Learning Systems}\/},  {\it \bibinfo{volume}{31}\/},
  \bibinfo{pages}{710--724}.
\bibitem[{Tran et~al.(2020)Tran, Kiranyaz, Gabbouj \&
  Iosifidis}]{tran2020progressive}
\bibinfo{author}{Tran, D.~T.}, \bibinfo{author}{Kiranyaz, S.},
  \bibinfo{author}{Gabbouj, M.}, \& \bibinfo{author}{Iosifidis, A.}
  (\bibinfo{year}{2020}).
\newblock \bibinfo{title}{Progressive operational perceptrons with memory}.
\newblock {\it \bibinfo{journal}{Neurocomputing}\/},  {\it
  \bibinfo{volume}{379}\/}, \bibinfo{pages}{172--181}.
\bibitem[{Tran et~al.(2017)Tran, Magris, Kanniainen, Gabbouj \&
  Iosifidis}]{Tran2017}
\bibinfo{author}{Tran, D.~T.}, \bibinfo{author}{Magris, M.},
  \bibinfo{author}{Kanniainen, J.}, \bibinfo{author}{Gabbouj, M.}, \&
  \bibinfo{author}{Iosifidis, A.} (\bibinfo{year}{2017}).
\newblock \bibinfo{title}{Tensor representation in high-frequency financial
  data for price change prediction}.
\newblock {\it \bibinfo{journal}{IEEE Symposium Series on Computational
  Intelligence}\/},  (pp. \bibinfo{pages}{1--7}).
\bibitem[{Tsantekidis et~al.(2017)Tsantekidis, Passalis, Tefas, Kanniainen,
  Gabbouj \& Iosifidis}]{tsantekidis2017forecasting}
\bibinfo{author}{Tsantekidis, A.}, \bibinfo{author}{Passalis, N.},
  \bibinfo{author}{Tefas, A.}, \bibinfo{author}{Kanniainen, J.},
  \bibinfo{author}{Gabbouj, M.}, \& \bibinfo{author}{Iosifidis, A.}
  (\bibinfo{year}{2017}).
\newblock \bibinfo{title}{Forecasting stock prices from the limit order book
  using convolutional neural networks}.
\newblock In {\it \bibinfo{booktitle}{IEEE Conference on Business
  Informatics}\/} (pp. \bibinfo{pages}{7--12}).
\newblock volume~\bibinfo{volume}{1}.
\bibitem[{Tsantekidis et~al.(2020)Tsantekidis, Passalis, Tefas, Kanniainen,
  Gabbouj \& Iosifidis}]{Tsantekidis2020}
\bibinfo{author}{Tsantekidis, A.}, \bibinfo{author}{Passalis, N.},
  \bibinfo{author}{Tefas, A.}, \bibinfo{author}{Kanniainen, J.},
  \bibinfo{author}{Gabbouj, M.}, \& \bibinfo{author}{Iosifidis, A.}
  (\bibinfo{year}{2020}).
\newblock \bibinfo{title}{Using deep learning for price prediction by
  exploiting stationary limit order book features}.
\newblock {\it \bibinfo{journal}{Applied Soft Computing}\/},  {\it
  \bibinfo{volume}{93, 106401}\/}.
\bibitem[{Wang \& Deng(2018)}]{wang2018deep}
\bibinfo{author}{Wang, M.}, \& \bibinfo{author}{Deng, W.}
  (\bibinfo{year}{2018}).
\newblock \bibinfo{title}{Deep visual domain adaptation: A survey}.
\newblock {\it \bibinfo{journal}{Neurocomputing}\/},  {\it
  \bibinfo{volume}{312}\/}, \bibinfo{pages}{135--153}.
\bibitem[{Wang \& Han(2014)}]{wang2014online}
\bibinfo{author}{Wang, X.}, \& \bibinfo{author}{Han, M.}
  (\bibinfo{year}{2014}).
\newblock \bibinfo{title}{Online sequential extreme learning machine with
  kernels for nonstationary time series prediction}.
\newblock {\it \bibinfo{journal}{Neurocomputing}\/},  {\it
  \bibinfo{volume}{145}\/}, \bibinfo{pages}{90--97}.
\bibitem[{Wang et~al.(2019)Wang, Du \& Guo}]{wang2019domain}
\bibinfo{author}{Wang, Z.}, \bibinfo{author}{Du, B.}, \& \bibinfo{author}{Guo,
  Y.} (\bibinfo{year}{2019}).
\newblock \bibinfo{title}{Domain adaptation with neural embedding matching}.
\newblock {\it \bibinfo{journal}{IEEE Transactions on Neural Networks and
  Learning Systems}\/},  {\it \bibinfo{volume}{31}\/},
  \bibinfo{pages}{2387--2397}.
\bibitem[{Yang et~al.(2015)Yang, Jing, Yu \& Ng}]{yang2015learning}
\bibinfo{author}{Yang, L.}, \bibinfo{author}{Jing, L.}, \bibinfo{author}{Yu,
  J.}, \& \bibinfo{author}{Ng, M.~K.} (\bibinfo{year}{2015}).
\newblock \bibinfo{title}{Learning transferred weights from co-occurrence data
  for heterogeneous transfer learning}.
\newblock {\it \bibinfo{journal}{IEEE Transactions on Neural Networks and
  Learning Systems}\/},  {\it \bibinfo{volume}{27}\/},
  \bibinfo{pages}{2187--2200}.
\bibitem[{Ye \& Dai(2018)}]{ye2018novel}
\bibinfo{author}{Ye, R.}, \& \bibinfo{author}{Dai, Q.} (\bibinfo{year}{2018}).
\newblock \bibinfo{title}{A novel transfer learning framework for time series
  forecasting}.
\newblock {\it \bibinfo{journal}{Knowledge-Based Systems}\/},  {\it
  \bibinfo{volume}{156}\/}, \bibinfo{pages}{74--99}.
\bibitem[{Yu et~al.(2007)Yu, Wang \& Lai}]{yu2007online}
\bibinfo{author}{Yu, L.}, \bibinfo{author}{Wang, S.}, \& \bibinfo{author}{Lai,
  K.~K.} (\bibinfo{year}{2007}).
\newblock \bibinfo{title}{An online learning algorithm with adaptive forgetting
  factors for feedforward neural networks in financial time series
  forecasting}.
\newblock {\it \bibinfo{journal}{Nonlinear dynamics and systems theory}\/},
  {\it \bibinfo{volume}{7}\/}, \bibinfo{pages}{51--66}.
\bibitem[{Zhang et~al.(2019)Zhang, Zohren \& Roberts}]{Zhang2019}
\bibinfo{author}{Zhang, Z.}, \bibinfo{author}{Zohren, S.}, \&
  \bibinfo{author}{Roberts, S.} (\bibinfo{year}{2019}).
\newblock \bibinfo{title}{Deeplob: Deep convolutional neural networks for limit
  order books}.
\newblock {\it \bibinfo{journal}{IEEE Transactions on Signal Processing}\/},
  {\it \bibinfo{volume}{67}\/}, \bibinfo{pages}{3001--3012}.
\bibitem[{Zhao et~al.(2014)Zhao, Hoi, Wang \& Li}]{zhao2014online}
\bibinfo{author}{Zhao, P.}, \bibinfo{author}{Hoi, S.~C.},
  \bibinfo{author}{Wang, J.}, \& \bibinfo{author}{Li, B.}
  (\bibinfo{year}{2014}).
\newblock \bibinfo{title}{Online transfer learning}.
\newblock {\it \bibinfo{journal}{Artificial Intelligence}\/},  {\it
  \bibinfo{volume}{216}\/}, \bibinfo{pages}{76--102}.
\bibitem[{Zhao et~al.(2020)Zhao, Yue, Zhang, Li, Zhao, Wu, Krishna, Gonzalez,
  Sangiovanni-Vincentelli, Seshia et~al.}]{zhao2020review}
\bibinfo{author}{Zhao, S.}, \bibinfo{author}{Yue, X.}, \bibinfo{author}{Zhang,
  S.}, \bibinfo{author}{Li, B.}, \bibinfo{author}{Zhao, H.},
  \bibinfo{author}{Wu, B.}, \bibinfo{author}{Krishna, R.},
  \bibinfo{author}{Gonzalez, J.~E.}, \bibinfo{author}{Sangiovanni-Vincentelli,
  A.~L.}, \bibinfo{author}{Seshia, S.~A.} et~al. (\bibinfo{year}{2020}).
\newblock \bibinfo{title}{A review of single-source deep unsupervised visual
  domain adaptation}.
\newblock {\it \bibinfo{journal}{IEEE Transactions on Neural Networks and
  Learning Systems}\/},  {\it \bibinfo{volume}{33}\/},
  \bibinfo{pages}{473--493}.
\bibitem[{Zuo et~al.(2016)Zuo, Zhang, Pedrycz, Behbood \& Lu}]{zuo2016fuzzy}
\bibinfo{author}{Zuo, H.}, \bibinfo{author}{Zhang, G.},
  \bibinfo{author}{Pedrycz, W.}, \bibinfo{author}{Behbood, V.}, \&
  \bibinfo{author}{Lu, J.} (\bibinfo{year}{2016}).
\newblock \bibinfo{title}{Fuzzy regression transfer learning in takagi--sugeno
  fuzzy models}.
\newblock {\it \bibinfo{journal}{IEEE Transactions on Fuzzy Systems}\/},  {\it
  \bibinfo{volume}{25}\/}, \bibinfo{pages}{1795--1807}.

\end{thebibliography}

\end{document}